\documentclass{article}
\usepackage{arxiv}
\usepackage[utf8]{inputenc} 
\usepackage[T1]{fontenc}    
\usepackage{hyperref}       
\usepackage{url}            
\usepackage{booktabs}       
\usepackage{amsmath,amsfonts}       
\usepackage{nicefrac}       
\usepackage{microtype}      
\usepackage{lipsum}		
\usepackage{graphicx}
\usepackage{natbib}
\usepackage{tabularx}
\usepackage{longtable}
\usepackage{doi}
\usepackage{algorithm}
\usepackage{algpseudocode}
\usepackage{array}
\usepackage{amssymb}
\usepackage{natbib}

\newcolumntype{R}[1]{>{\raggedleft\arraybackslash}m{#1}}  

\newcolumntype{L}[1]{>{\raggedright\arraybackslash}m{#1}}

\newcolumntype{C}[1]{>{\centering\arraybackslash}m{#1}}

\title{A Review of Neural Networks in Precipitation Prediction}

\author{ \href{https://orcid.org/0009-0005-1489-5911}{\includegraphics[scale=0.06]{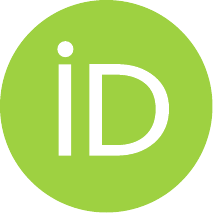}\hspace{1mm}Yugong Zeng}\\
	Department of Electrical and Computer Engineering\\
	University of Windsor\\
    401 Sunset Avenue\\
	Ontario, Canada N9B 3P4 \\
	\texttt{zeng26@uwindsor.ca} \\
    \And
	{\hspace{1mm}Jiayuan Wang} \\
	Department of Electrical and Computer Engineering\\
	University of Windsor\\
	401 Sunset Avenue \\
    Ontario, Canada N9B 3P4 \\
	\texttt{wang621@uwindsor.ca} \\
	\And
	{\hspace{1mm}Jonathan Wu} \\
	Department of Electrical and Computer Engineering\\
	University of Windsor\\
	401 Sunset Avenue \\
    Ontario, Canada N9B 3P4 \\
	\texttt{jwu@uwindsor.ca} \\
}

\date{}

\renewcommand{\shorttitle}{\textit{}}

\begin{document}
\maketitle

\begin{abstract}
Precipitation prediction has undergone a profound transformation. A notable limitation of traditional NWP is the need for extensive statistical post-processing. To address this challenge, neural network–based approaches were developed. These approaches offer a framework that directly learns the mapping from atmospheric predictors to precipitation targets. Based on the technological development, this article first reviews the traditional precipitation forecasting methods and summarizes the development trends of precipitation forecasting based on neural networks. We then outline the training process, loss functions, and some datasets for precipitation prediction. In the main body of the article, we detail the basic artificial neural networks (ANNs), spatial feature extraction models, time feature extraction models, generative models, Transformer models, graph neural networks (GNNs), and emerging hybrid models. Finally, in the appendix, we supplement the commonly used evaluation metrics. This paper focuses on the advantages and disadvantages of various neural network models in precipitation forecasting applications, and also pays attention to the latest progress of neural network-based methods. Overall, neural networks have significantly improved the accuracy of short-term and medium-term precipitation forecasting, but still face challenges in representing extreme rainfall, handling imbalanced data, and ensuring physical consistency. The latest progress shows that future prediction systems will increasingly rely on the integration of multiple sources of data and hybrid physical-data-driven models to enhance their robustness and applicability. By compositing research covering multiple eras and paradigms, we not only depict the history of neural networks in precipitation prediction but also outline future directions in next generation forecasting systems.
\end{abstract}


\section{Introduction}
Precipitation prediction is the cornerstone of modern environmental forecasting, laying the foundation for research and decision-making in fields such as agriculture, disaster management, hydrology, transportation, and climate. More specifically, accurate rainfall forecasting is crucial for reducing the risks of floods and droughts, managing water resources, harvesting crops, and providing early warnings of extreme weather.  In fields such as aviation and ground transportation, precipitation forecasts play a key role in short-term operational decisions as well as long-term strategic planning. Moreover, according to the view of scholars in the field of climate change, understanding precipitation patterns directly contributes to assessing ecological risks, and the sustainability of nature\cite{Li2022}. For these reasons, improving the accuracy and scope of precipitation forecasting has become not only a scientific research direction but also a policy priority.

Since its emergence as a formal scientific discipline in the 19th century, geophysics has steadily evolved from the study of Earth's physical properties to a multifaceted field encompassing climate dynamics, natural hazard assessment, and hydrological modeling\cite{Historyofgeophysics}. One of the most influential applications is precipitation prediction, which integrates knowledge from multiple disciplines such as atmospheric physics, environmental monitoring, and computational modeling. In the 1940s, the idea of a universal computing machine appeared. Later, programmable digital computers and high-level programming languages were developed. Computers then became practical tools for automating complex analysis. One of the impactful applications was NWP, where pre-existing atmospheric equations could finally be rapidly discretized and solved using Finite Difference Methods in early machines. Although early NWP systems were groundbreaking, the workflow was largely designed by humans. It typically began with manual data initialization and objective analysis, followed by variable selection. Then the model was formulated and the forecast was generated in a structured pipeline.

As computer science advanced, meteorological observation technology also made significant breakthroughs. After the application of remote sensing technology shifted from military use to civilian use, the observation and recording of meteorological phenomena have become more detailed. The launch of satellites in the latter half of the 20th century marked a major turning point, making it possible to conduct systematic and high-resolution monitoring of meteorological phenomena\cite{Anderson2010}. Scientists began to combine computer and remote sensing technologies to conduct global precipitation analysis.

Machine learning eventually revolutionized the forecasting domain. Although theoretical roots developed with early computers in the 1950s, the field matured in the 1990s. In recent years, machine learning has developed toward the end-to-end learning paradigm, where models can learn to directly map raw observational data to final rainfall prediction. These models are typically constructed based on neural networks and are trained throughout the process, thereby achieving higher levels of automation and prediction performance\cite{1}.

The main part of this review traces the evolution of neural networks in precipitation prediction, from simple ANNs to hybrid neural networks. By reviewing emerging research directions, our aim is to comprehensively present the evolution process of precipitation prediction under different modeling approaches.

\section{Literature Survey}
\subsection{Basics of Precipitation Prediction}
Precipitation prediction refers to estimating the occurrence, type, intensity and spatial distribution of drizzle, rainfall, snowfall or other products of atmospheric water vapor over a given forecast horizon. Founded in the disciplines of meteorology, atmospheric physics and hydrology, it combines physical modeling of atmospheric dynamics with observational data analysis to anticipate the evolution of weather systems. This process is inherently multiscale, involving interactions across microphysical processes, mesoscale dynamics and large-scale climate drivers such as the El~Ni\~{n}o--Southern Oscillation (ENSO)\cite{glantz2001currents}, \cite{Trenberth2004}.

\subsection{Traditional Predictive Methods}Centuries ago, meteorological prediction was largely imprecise and relied heavily on intuition, local climatology and empirical rules. Observational data were sparse, especially over oceans and in the upper atmosphere, and theoretical physics played little role in practice. Predictions were produced by plotting surface observations on weather maps and manually identifying patterns\cite{Harper2024}.

With the development of theoretical physics and astronomy, in the last decade of 19th century, early pioneers such as Cleveland Abbe and Vilhelm Bjerknes illustrated the foundational principles of scientific weather prediction. Vilhelm Bjerknes also proposed a systematic framework involving diagnostic and prognostic steps based on governing equations of the atmosphere\cite{Friedman1989}. This vision was carried forward by Lewis Fry Richardson, who attempted the first numerical weather forecast using finite difference approximations. Although his effort resulted in unrealistic outcomes due to data imbalance and the lack of computational resources, Richardson still introduced the core concept of modeling atmospheric evolution via numerical integration of physical equations\cite{LYNCH20083431}.

Modern meteorology changed to a deterministic framework as the governing equations of fluid motion and thermodynamics were formalized into NWP. The first NWP utilized the barotropic vorticity equation, treating the atmosphere as a single and uniform layer\cite{Charney01011950}. Later, many multi-level primitive equation models were introduced into operations. But early NWP systems struggled with short-term and highly localized forecasts, so researchers developed statistical post-processing methods to improve it\cite{Murphy1988}. In 1972, Harry R. Glahn and Dale A. Lowryit introduced Model Output Statistics (MOS) with screening regression into weather forecasting, specifically, Probability of Precipitation (PoP)\cite{Harry1972}. By linearly combining a series of independent predictors, named $X_0$ to $X_k$, they denoted a predictand $\hat{Y}$ as:
\begin{equation*}
\hat{Y}=a_0+a_1X_1+a_2X_2+...+a_kX_k,
\end{equation*}
where $a_i$ is called the coefficient. Then, they applied root mean square error (RMSE) on dependent samples of size $n$ to determine the goodness of the equation for estimation. By leveraging screening regression, which iteratively selects a parsimonious subset minimizing estimated error, they derived a statistically optimized equation that translated NWP-derived predictors. This method significantly outperformed climatological and persistence-based forecasts, and also addressed key limitations of raw model outputs by accounting for systematic biases and local climatological characteristics. In 1990, Jan John and Josef \v{S}tekl expanded this method into nonlinear adaptive regression predictors, based on singular value decomposition (SVD)\cite{John1990}. In their setup, SVD acted as a stabilizer and a smart filter. It automatically removed useless and repetitive data while keeping the meaningful patterns. Since adaptive regression for temperature showed promise, SVD was soon applied to precipitation prediction. For example, Liu developed a forecast model by applying a lagged SVD between soil moisture fields and subsequent monthly--seasonal precipitation\cite{Liu2003}. This model effectively predicted regional precipitation, highlighting the practicality of SVD in dynamic information-driven rainfall prediction.

In the 21st century, the scope and complexity of global precipitation forecast systems expanded further. The National Oceanic and Atmospheric Administration (NOAA) had previously relied on separate models for different forecast ranges, namely the Medium Range Forecast (MRF) and Aviation (AVN) models. These models were unified into the Global Forecast System (GFS) in 2002 \cite{White2018GFSHistory}\cite{Tracton1993}. The GFS provided continuous forecasts up to 16 days and incorporated various Earth system components, including land surface, sea ice, and radiation processes. In the same year, the German Weather Service (DWD) implemented the Global Model (GME), becoming the first agency to adopt an icosahedral--hexagonal grid, and enhancing precipitation prediction in polar regions by removing grid singularities\cite{Detlev2002}. NOAA developed the finite-volume icosahedral model (FIM) in the late 2000s and early 2010s as the core solution for its next-generation global dynamic model, aiming to enhance the scalability and resolution of the operational GFS\cite{Lee2006}\cite{Henderson2010}. 

In recent years, large-scale operational NWP systems have improved in accuracy, and thus have been more widely applied in precipitation prediction. For example, the Integrated Forecasting System (IFS), which was developed by the European Centre for Medium-Range Weather Forecasts (ECMWF), represents one of the most influential global forecasting frameworks and frequently serves as a benchmark for evaluating emerging neural network-based rainfall forecasting methods\cite{wedi2015modelling}. Similarly, the Global/Regional Assimilation and Prediction System (GRAPES), which was developed by the China Meteorological Administration, provides operational forecasting capabilities on both global and regional domains and is often applied for precipitation prediction in East Asia\cite{Chen2008}. 

However, NWP systems are still facing challenges such as the chaotic nature of the atmosphere, nonlinear relationships between variables, and limited observational resolution\cite{Kalnay_2002}. Errors can arise from uncertainties in initial conditions, boundary constraints, and model physics, particularly in convective-scale forecasting\cite{Zhang2003}. As a result, forecasting precision tends to decline with an increase in lead time and a decrease in spatial scale. To solve these challenges, recent studies have emphasized the use of data assimilation\cite{Houtekamer2016} and ensemble modeling\cite{Palmer_2000} to enhance model calibration and uncertainty quantification. These developments have gradually reshaped precipitation forecasting from a purely physics-based field into a hybridized discipline that balances theory, data, and computation.

\subsection{Trend of Recent Research}
\begin{figure}[!t]
\centering
\includegraphics[width=0.5\linewidth]{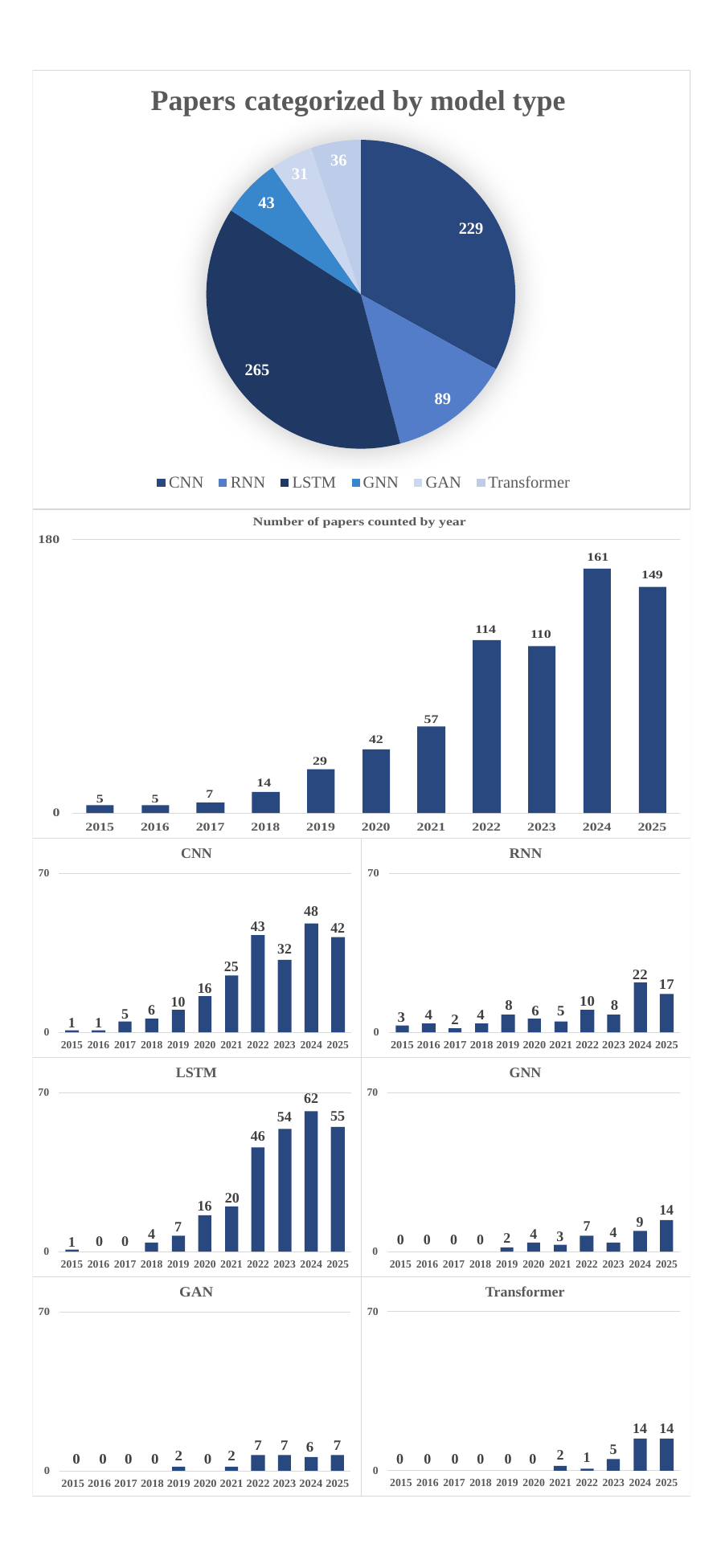}
\caption{Evolution and distribution of neural network–based precipitation prediction studies from January 2015 to October 2025, categorized by model architecture (CNN, RNN, LSTM, GNN, GAN, and Transformer).}
\label{papers}
\end{figure}

We collected papers related to this field published in the last 10 years from Google Scholar, and plotted their categories and amounts in Fig.~\ref{papers}. To ensure methodological transparency, the literature trend analysis was conducted using structured queries. To ensure methodological transparency, the literature trend analysis was conducted using structured queries. For instance, specific title searches were performed using combinations of keywords (e.g., convolutional OR CNN, precipitation OR rainfall, prediction OR forecast) in finding convolutional neural network–based precipitation prediction articles. To ensure the results were highly relevant, filters were applied to remove studies focused on precipitation estimation, classification, bias correction, calibration, statistical downscaling or runoff modeling. These exclusions were implemented using Boolean negation operators (e.g., '-estimating', '-downscaling', '-runoff'). In addition, to avoid redundancy (e.g., paper title including 'convolutional' and 'CNN'), overlapping matches were cross-checked, and redundant results were removed. Similar query logic was applied to other categories of neural networks. One point worth mentioning is that since there have been few articles on ANNs and deep neural networks (DNNs) in the past 10 years, we excluded them when conducting the statistics. While Google Scholar does not guarantee exhaustive coverage, this strategy provides a broad and representative overview of research trends in neural network–based precipitation prediction. As illustrated in Fig.~\ref{papers}, we analyzed the evolution and distribution of precipitation prediction research using neural network architectures between January 2015 and October 2025. We identified a total of 628 papers during this decade, indicating that neural networks have been widely applied. Moreover, the annual publication volume of research articles has gradually increased from a few in the early years to over a hundred in recent years, suggesting that the application of neural networks in precipitation prediction is becoming increasingly widespread. Among the surveyed architectures, long short-term memory (LSTM) networks and convolutional neural networks (CNNs) dominate the literature, accounting for approximately 42\% and 36\% of total publications, respectively. CNN-based models have become popular since 2019. As spatial representation learning becomes increasingly crucial for radar echo and satellite image processing, their usage rate has remained at a relatively high level. Similarly, LSTMs witnessed a rapid expansion after 2021. Because of their ability to capture complex temporal dependencies, LSTMs were rapidly adopted as a mainstream solution for precipitation prediction. Recurrent neural networks (RNNs), an early sequence modeling method, held a central position from 2015 to 2016. However, with the emergence of more advanced architectures LSTMs and Transformers, although they still maintain a certain level of application, their proportion in the overall number of published papers has actually decreased. GNN was first applied to precipitation prediction around 2019. In recent years, due to its ability to simulate the irregular spatial dependencies between meteorological stations and grid networks, it has developed rapidly. Generative adversarial networks (GANs) and Transformers, although they emerged later, have also experienced exponential growth since 2022. GANs are increasingly used for high-resolution rainfall field generation and spatial downscaling, while Transformers demonstrate superior capability in capturing long-range dependencies, making them promising candidates for next-generation weather forecasting systems.

\subsection{Problem Presentation}
Precipitation prediction is fundamentally a spatiotemporal estimation problem, where the objective is to learn the nonlinear mapping between historical atmospheric parameters and future precipitation states\cite{An2025}. In this review, we follow the scope of precipitation prediction, distinguishing from precipitation retrieval, which aims to estimate rainfall rates at the observation time based on satellite or radar measurements and is typically formulated as an inverse problem\cite{PRIGENT2010380}. In contrast, precipitation prediction seeks to forecast future precipitation states, given historical time-series atmospheric observations or model outputs\cite{KUNDU2023}\cite{Poornima2019}.
\subsubsection{Training Process}
Traditionally, precipitation is computed by moisture advection, for example, in NWP model, physical parameterization equations are used to simulate precipitation formation processes for temperatures above 0$\,^\circ\mathrm{C}$.
Unlike traditional empirical or physics-based methods, neural networks provide a data-driven framework capable of approximating such mappings without relying on explicit physical parameterizations\cite{Du2016}. Specifically, neural networks automatically extract multi-scale representations from input data such as radar reflectivity, satellite infrared (IR) imagery, or reanalysis fields. The core theoretical concept of neural networks is to minimize the error between the predicted precipitation $\hat{Y}$ and the observed value $Y$. Generally, precipitation prediction can be formulated as a spatiotemporal forecasting problem that seeks to learn a nonlinear mapping from historical atmospheric states to future precipitation fields. Let $X_{t-k:t}$ denote a sequence of past atmospheric variables (e.g., radar reflectivity, satellite imagery, reanalysis fields), and let $Y_{t+\Delta t}$ denote the target precipitation field at lead time $\Delta t$. The objective is to approximate a function:
\[
\mathcal{F}: X_{t-k:t} \rightarrow Y_{t+\Delta t},
\]
where $\mathcal{F}$ represents a neural network parameterized by $\theta$. In practice, neural networks learn this mapping by automatically extracting hierarchical, multi-scale representations from the input data. Then, parameters $\theta$ are optimized by minimizing the error that measures the discrepancy between predictions and observations. Although architectures vary widely, the training of neural networks for precipitation prediction follows a structured sequence of operations that transforms raw meteorological inputs into optimized parameters capable of producing accurate precipitation forecasts.

The first stage is forward propagation, which applies a series of linear and nonlinear transformations parameterized by $\theta = \{W, b\}$, where $W$ denotes the trainable weights and $b$ the bias terms. Each layer computes an output as follows:
\[
\mathbf{h}^{(l)} = f^{(l)}(W^{(l)} \mathbf{h}^{(l-1)} + b^{(l)}),
\]
where $l$ is the layer index, $\mathbf{h}^{(l)}$ and $f^{(l)}$ denote the extracted meteorological features and the applied activation function in layer $l$, respectively. In ANNs and DNNs, $W^{(l)}$ represents dense weight matrices that map the learned representations from one layer to the next, enabling hierarchical abstraction of atmospheric patterns\cite{Shamshirband2015}. In convolutional architectures, $W^{(l)}$ corresponds to learnable parameters of the convolutional layer that extract spatial features such as cloud edges or rainband structures. For recurrent or transformer-based networks, the operation extends across temporal or relational dependencies to capture dynamic dependencies in the rainfall evolution\cite{Inoue2022}.

The second stage is loss computation. Network's output is denoted as $\hat{Y}_{t+\Delta t}$, where $\Delta t$ is based on the target of the predictive task\cite{Tan2024}\cite{Weesakul2018}). $\hat{Y}_{t+\Delta t}$ is compared with the ground-truth precipitation $Y_{t+\Delta t}$, using a task-specific loss function $\mathcal{L}(\theta)$. The loss quantifies the model's error and serves as the feedback signal for learning, and can be written as: 
\[
\mathcal{L}(\theta)=\frac{1}{N}\sum_{n=1}^{N}\mathcal{D}\!\left(\hat{Y}^{(n)}_{t+\Delta t},\,Y^{(n)}_{t+\Delta t}\right)
\]
where $N$ denotes the total number of samples, $n$ is the sample index, and $\mathcal{D}(\cdot)$ is a distance or divergence metric. Once the loss is computed, its gradient with respect to $i$-th trainable parameter $\theta_i$ is obtained via backpropagation:
\[
\frac{\partial \mathcal{L}}{\partial \theta_i} = \frac{\partial \mathcal{L}}{\partial \hat{Y}_{t+\Delta t}} \cdot \frac{\partial \hat{Y}_{t+\Delta t}}{\partial \theta_i}.
\]
These gradients indicate the direction and magnitude by which parameters should be adjusted to minimize prediction error. To perform this adjustment, an optimization algorithm (e.g., stochastic gradient descent, Adam, RMSProp\cite{Zaheer2019}) updates the parameters at iteration $k$ as follows:
\[
\theta^{(k+1)} = \theta^{(k)} - \eta \, \nabla_\theta \mathcal{L}(\theta^{(k)}),
\]
where $\eta$ represents the learning rate. The process continues until the loss converges or a stopping criterion (such as early stopping or validation plateau) is met.

Although the above optimization process is quite common in neural networks, there are some unique challenges in the application of precipitation prediction. For instance, when a network trained in an area with dense observational data is transferred to an area with sparse observational data, the predictive ability of the neural network often drops sharply. In addition, the distribution of precipitation also significantly affects the model training process. Training data contains extreme precipitation or no precipitation for a long time, and the incorrect processing of such values often leads to gradient disappearance or accuracy degradation. In the later sections, we will discuss and conclude how these specific neural networks solve the challenges.

\subsubsection{Loss Functions}
Beyond the training process, loss functions also play a crucial role in precipitation prediction because rainfall distributions are highly skewed\cite{Alcantara2020}. Standard regression losses such as mean squared error (MSE), mean absolute error (MAE), and their combination \cite{PREDRNNwang18}, remain widely used due to their stable optimization behavior:
\[
\mathcal{L}_{\mathrm{MSE}} = \frac{1}{N} \sum_{i=1}^{N} (\hat{y}_i - y_i)^2,
\quad
\mathcal{L}_{\mathrm{MAE}} = \frac{1}{N} \sum_{i=1}^{N} |\hat{y}_i - y_i|,
\]
where $\hat{y}_i$ and $y_i$ denote the predicted and observed precipitation of the index $i$, respectively. However, these standard losses are often dominated by high-frequency but low-intensity samples, which systematically bias the model toward underestimating rare but high-impact precipitation events. To address this class imbalance, weighted loss functions are employed to assign higher importance to moderate and heavy rainfall regions\cite{shi2017deep}\cite{Franch2020}\cite{Hess2022}. For example, weighted MSE:
\[
\mathcal{L}_{\mathrm{wMSE}} = \frac{1}{N} \sum_{i=1}^{N} w(y_i)\,(\hat{y}_i - y_i)^2,
\]
where $w(\cdot)$ denotes an intensity-dependent weight that increases with rainfall magnitude, thereby encouraging the model to focus more on extreme precipitation. However, temporal imbalance also exists in predictive sequences: later predicted frames are typically harder due to error accumulation, yet standard MAE/MSE assigns an equal penalty to all frames. Ma et al. proposed focal frame loss (FFL), which adaptively reweights each predicted frame according to its relative prediction difficulty, yielding focal-frame variants of MAE/MSE (FF-MAE/FF-MSE) and improving multiple radar nowcasting models without extra training overhead\cite{Ma2022FFLOSS}. Similar cost-sensitive formulations are frequently adopted to reduce the dominance of non-rain samples during training. A practical alternative frequently adopted in precipitation modeling is to operate in logarithmic space, which stabilizes gradients in a gamma-like distribution of the rain amount\cite{Campos2023}:
\[
\mathcal{L}_{\text{MSLE}}
= \frac{1}{N} \sum_{i=1}^{N}
\left(
\log(1+\hat{y}_i) - \log(1+y_i)
\right)^2,
\]
Another logarithmic loss function, which combines the stability of mean squared error for small deviations with robustness to outliers, is named log-cosh loss\cite{Ayzel2020}:
\[
\mathcal{L}
= \frac{1}{N} \sum_{i=1}^{N}
\log\!\left(\cosh(\hat{y}_i - y_i)\right),
\quad
\cosh(x) = \frac{e^x + e^{-x}}{2}.
\]
Adversarial or generative training strategies have also become important in precipitation nowcasting to improve spatial realism. A typical objective combines reconstruction and adversarial terms is defined as follows\cite{YOU2023}:
\[
\mathcal{L} = \mathcal{L}_{\mathrm{rec}} + \lambda \mathcal{L}_{\mathrm{adv}},
\]
where $\mathcal{L}_{\mathrm{rec}}$ denotes the reconstruction loss, $\mathcal{L}_{\mathrm{adv}}$ penalizes unrealistic rainfall structures and reduces the spatial smoothing effects of pixel-wise losses, and $\lambda$ serves as a weighting hyperparameter to balance the two objectives.

Finally, recent studies have explored training objectives aligned with operational verification metrics such as the critical success index (CSI) or the false alarm ratio (FAR)\cite{KIM2024}. By integrating these metrics through differentiable surrogate losses, researchers can predict the initiation and evolution of heavy rainfall systems more reliably.

\subsection{Modern Technologies: Introduction of Datasets}
A key factor for the rapid development of neural networks is the increasing availability of large-scale, high-quality datasets which provide spatiotemporal records of atmospheric variables for training and benchmarking. Therefore, we concluded a series of the most commonly used datasets prior to introducing machine learning models.
\subsubsection{ERA5}
ERA5 (ECMWF Reanalysis v5), the fifth-generation global atmospheric reanalysis, produced by the Copernicus Climate Change Service (C3S) at ECMWF, spans from January 1940 to the present\cite{Soci2024}. It delivers hourly estimates of a wide array of atmospheric, land-surface, and ocean-wave variables on a global 31~km grid ($0.25^{\circ}$). Vertically, ERA5 resolves the atmosphere on 137 hybrid model levels up to 0.01~hPa, with additional products interpolated to 37 pressure levels. In addition to the high-resolution deterministic product (HRES), ERA5 includes a 10-member ensemble of data assimilations (EDA), available at reduced spatial (63~km) and temporal (3-hourly) resolution, which provides flow-dependent uncertainty estimates. The observing system assimilated in ERA5 has expanded from 17,000 reports per day in 1940 to 25 million per day in 2022, reflecting the incorporation of aircraft, balloon-borne instruments, buoys, ships, and satellite observations. Accordingly, the quality of the reanalysis improves substantially throughout the record. Recent refinements include improved treatment of tropical cyclone observations from the International Best Track Archive for Climate Stewardship (IBTrACS) and bias correction of surface pressure records, leading to a more homogeneous representation of extreme events and soil moisture consistency. 

In addition, ensemble mean and spread are directly accessible to quantify uncertainty. The dataset is updated in near real-time via the ERA5T product, which operates with a latency of approximately five days\cite{ERA5t2022}. These preliminary records are subsequently replaced by final, quality-controlled versions after two to three months, ensuring both timeliness and reliability for climate and forecasting applications. Although there is still uncertainty in the data from the first few decades, especially in the Southern Hemisphere where observational data is sparse, ERA5 provides a long-term, high-resolution, and physically consistent record of atmospheric variables. This makes it an ideal benchmark and a key training dataset for modern neural networks in precipitation prediction.

\subsubsection{SEAS5}
The ECMWF's fifth generation seasonal forecast system (SEAS5) has been in operation since November 2017 and aims to provide global climate forecasts ranging across subseasonal-to-seasonal (S2S) timescales\cite{Johnson2019}. In terms of atmospheric modeling, SEAS5 is based on ECMWF’s IFS, utilizing a model cycle closely aligned with operational medium-range forecasts but specifically configured for the extended integrations required of seasonal timescales. For ocean modeling, SEAS5 uses the Nucleus for European Modelling of the Ocean (NEMO) model.
Currently, SEAS5 issues ensemble forecasts with 51 members initialized on the first day of each month. While each forecast typically extends 7 months ahead, a subset of 15 members is extended once per quarter to run for 13 months. This ensemble framework captures forecast uncertainty and allows for robust probabilistic prediction. Horizontally, the atmospheric grid spacing of SEAS5 is approximately 36~km, and the ocean resolution is $0.25^\circ$. In the vertical direction, the atmospheric model has 91 layers, with the highest layer reaching the mesosphere of 0.01 hPa (about 80 km), and the ocean model has 75 layers. Temporally, the system provides forecast outputs at 6-hourly and daily intervals\cite{Hylke2022}. The system assimilates ocean and atmosphere observations for initialization, and it includes interactive sea ice and land surface components.

Relative to its predecessor (System 4), SEAS5 introduced improved ENSO prediction skill, more realistic sea-ice initialization, better representation of stratospheric processes and upgraded ocean physics. These improvements enhance forecast reliability for precipitation, temperature anomalies, and extreme-event probabilities on seasonal timescales. SEAS5 outputs, including ensemble-based forecasts of precipitation anomalies, are widely used to support seasonal precipitation prediction in hydrology, agriculture, energy and disaster risk management, and are publicly available via the Copernicus Climate Data Store.

\subsubsection{GOES Satellite Imagery Dataset}
The Geostationary Operational Environmental Satellites (GOES) program, operated by NOAA in partnership with NASA, provides continuous geostationary observations of the Western Hemisphere. Since the launch of the GOES-R series (GOES-16 in 2016, GOES-17 in 2018 and GOES-18 in 2022)\cite{GOODMAN2020}\cite{Lotoaniu2023}, the primary instrument, the Advanced Baseline Imager (ABI), has delivered high-resolution multispectral imagery across 16 spectral bands, including visible, IR, and near-IR channels\cite{Timothy2017}.

GOES ABI imagery is produced at spatial resolutions of 0.5~km (visible), 1~km (near-IR), and 2~km (IR bands), with temporal refresh rates of 5--10 minutes for full-disk coverage, 15 minutes for hemispheric scans, and as frequent as 30--60 seconds in mesoscale sectors. Among these, the IR channels (10--12 $\mu$m) are especially important for precipitation-related applications, as they provide cloud-top brightness temperature information that correlates with convective activity.

While GOES imagery itself represents radiance and brightness temperature fields, it serves as the basis for numerous precipitation estimation algorithms and products. Examples include the Hydro-Estimator (HE)\cite{Roderick2003}\cite{Yucel2011}, the GOES Rainfall Rate/Quantitative Precipitation Estimate (RRQPE)\cite{Kuligowski2013}\cite{Vesta2023}, and multi-sensor merged products such as Integrated Multi-satellitE Retrievals for GPM (IMERG)\cite{Huffman2020}, which incorporates GOES IR data to improve temporal sampling.

Therefore, GOES imagery is widely used in short-term forecasting, extreme weather monitoring, and as an input feature for machine learning models (for precipitation prediction). Its combination of high spatial resolution, rapid refresh rates and long-term continuity (dating back to the first image of Earth taken by GOES-1 in 1975) makes it a cornerstone observational dataset for precipitation science in the Americas.

\subsubsection{PRISM}
The Parameter-elevation Regressions on Independent Slopes Model (PRISM) dataset, developed by Oregon State University, provides high-resolution gridded climate data for the conterminous United States\cite{Daly2021}. PRISM datasets cover multiple variables, with precipitation (rain and melted snow) as a primary element. PRISM precipitation products are available at a spatial resolution of 4~km, with multiple temporal aggregations, including daily, monthly, seasonal, and annual means, extending back to 1895 for long-term climate monitoring. To construct these detailed precipitation fields, the model collects physical gauge data from diverse sources. The uniqueness of PRISM lies in its adoption of climatologically aided interpolation. This technology carefully combines short-term weather records with long-term historical averages to reduce measurement bias and ensure temporal consistency. Furthermore, since 2002, PRISM has actively blended radar data into its interpolation framework to enhance accuracy for the central and eastern United States. 

PRISM datasets are widely validated and recognized for their superior performance in complex terrain, especially in the western U.S., where orographic gradients are pronounced. As such, PRISM is often considered the “climate dataset of record” for U.S. agencies, supporting applications in hydrology, ecology, agriculture, and climate change assessment.

For precipitation prediction research, PRISM serves a dual role. Firstly, it serves as a benchmark reference dataset for validating machine learning models and downscaling approaches. Its second role is an input training dataset which provides long, consistent and high-resolution precipitation records across diverse physiographic regimes.

\subsubsection{IMERG}
IMERG was developed jointly by NASA and JAXA under the GPM mission and provides a globally consistent, high-resolution precipitation dataset\cite{Huffman2020}. IMERG merges observations from a constellation of passive microwave sensors aboard multiple satellites, calibrated with the GPM Core Observatory’s Dual-Frequency Precipitation Radar (DPR) and Microwave Imager (GMI). IR observations are also incorporated to enhance temporal sampling. IMERG provides precipitation estimates on a $0.1^{\circ} \times 0.1^{\circ}$ latitude--longitude grid (10~km) at 30-minute temporal resolution, with global coverage between $60^{\circ}$N-$60^{\circ}$S (and partially outside of that latitude band), Data are released in three latency modes: Early Run (4-hour delay), Late Run (14-hour delay), and Final Run (3.5-month delay). The dataset also includes uncertainty estimates, probability fields, and quality indices.

IMERG data have been extensively validated against rain gauge networks worldwide and demonstrate robust skill across a range of climatic regimes. However, its performance is subject to known physical constraints. Furthermore, while the product captures large-scale patterns effectively, it often struggles to fully resolve the magnitude of localized extreme events. Despite these limitations, IMERG’s fine spatiotemporal resolution and near-real-time availability make it a cornerstone for training neural networks. 

\subsubsection{SEVIR}
The Storm EVent ImagRy (SEVIR) dataset is a large-scale, curated benchmark for applying deep learning to meteorological imaging tasks such as precipitation nowcasting, synthetic radar generation and storm classification\cite{Veillette2020}. SEVIR aims to integrate multiple sensing modalities into a spatiotemporally aligned dataset designed to accelerate research in weather and climate prediction. SEVIR consists of image sequences over 10,000 weather events across the contiguous United States (CONUS), each covering a 384~km $\times$ 384~km patch and a 4-hour temporal window, and each sequence is sampled with a 5-minute step. The dataset includes five complementary data types: three channels (i.e., C02, C09, C13) from the GOES-16 ABI, vertically integrated liquid (VIL) radar mosaics from the Next-Generation Radar (NEXRAD) and lightning flashes from the GOES-16 Geostationary Lightning Mapper (GLM). While these sensors possess varying native spatial resolutions ranging from 0.5 km to 2 km, the dataset provides them in a synchronized format as the 1 km VIL band.
This multimodal alignment enables studies of cross-sensor relationships between cloud dynamics, radiative properties, precipitation structure and convective activity. The dataset’s design addresses the major challenges of using raw GOES and NEXRAD archives, i.e., data volume, alignment complexity, and the scarcity of machine-learning-ready benchmarks. SEVIR is stored in HDF5 format and includes detailed metadata and an event catalog, which makes it easy to reproduce model training and evaluation. It has since become one of the most widely adopted resources for benchmarking deep neural architectures, especially in precipitation nowcasting and spatiotemporal generative forecasting.

\begin{table*}[t]
\caption{Comparison of commonly used datasets in precipitation prediction studies}
\label{dataset_comparison}
\centering
\small
\renewcommand{\arraystretch}{1.25}
\setlength\tabcolsep{1.2pt} 
\begin{tabular}{%
L{0.10\textwidth}
L{0.15\textwidth}
L{0.15\textwidth}
L{0.10\textwidth}
L{0.23\textwidth}
L{0.23\textwidth}}
\hline
\textbf{Dataset} & \textbf{Data Type} & \textbf{Spatial / Temporal Resolution} & \textbf{Coverage} & \textbf{Strengths} & \textbf{Limitations} \\
\hline
ERA5 & Reanalysis & $\sim$31 km / Hourly & Global & Physically consistent multi-variable fields; long-term record & Extreme precipitation magnitudes are likely underestimated \\
\hline
SEAS5 & Seasonal forecast & $\sim$36 km / Monthly forecasts & Global & Seasonal forecasting capability; ensemble outputs & Coarse resolution; limited for local extremes \\
\hline
GOES ABI & Geostationary satellite imagery & 0.5–2 km / 5–15 min  & Americas & High temporal resolution & Provides radiance, not direct rainfall \\
\hline
IMERG & Satellite precipitation product & 0.1° ($\sim$10 km) / 30 min & Nearly global (60°N–60°S) & Nearly global precipitation coverage; multi-sensor fusion & Retrieval uncertainty over complex terrain and snowfall regions \\
\hline
PRISM & Gauge-based interpolation & $\sim$4 km / Daily, Monthly, Seasonal, Annual & Conterminous U.S. & High-quality climatological precipitation over complex terrain & Limited geographic coverage \\
\hline
SEVIR & Radar + Satellite dataset & 1 km equivalent / 5 min & CONUS & High-resolution spatiotemporal storm evolution data & Regional coverage only \\
\hline
\end{tabular}
\end{table*}

Precipitation datasets differ in observation source, spatiotemporal resolution, and suitability for forecasting tasks. Reanalysis datasets such as ERA5, which is often used as large-scale atmospheric predictors in medium- and long-term forecasting models, provide dynamically consistent multivariate inputs for neural networks used to predict precipitation evolution\cite{Rasp2020}\cite{XU2025}\cite{JIANG2024}. However, evaluation studies indicate that ERA5 generally reproduces large-scale precipitation patterns but often underestimates the intensity of extreme daily rainfall \cite{Lavers2022}\cite{Rivoire2021}. In contrast, satellite-derived precipitation products such as IMERG are widely adopted in short-term forecasting due to their high resolution and multi-sensor data fusion\cite{GaoYanbo2022}, although retrieval uncertainty remains a challenge over complex terrain and mixed-phase precipitation regions\cite{ShenZH2024}. Radar-based resources, including SEVIR dataset and regional radar composites, are typically adopted in deep learning frameworks that require locally high spatiotemporal resolution to capture storm evolution\cite{LiDawei2025}\cite{Yang2024Motion}. Notably, many recent studies use multimodal data, such as ERA5 atmospheric predictors with radar or satellite precipitation observations, to balance large-scale physical consistency and fine-scale precipitation realism in neural network-based forecasting systems\cite{KIM2024}\cite{curcio2025}. Finally, a comparative summary of these datasets is provided in Table~\ref{dataset_comparison}.

\subsection{Single Neural Networks}
\subsubsection{ANN and DNN Based Models}
\begin{figure}
\centering
\includegraphics[width=1\linewidth]{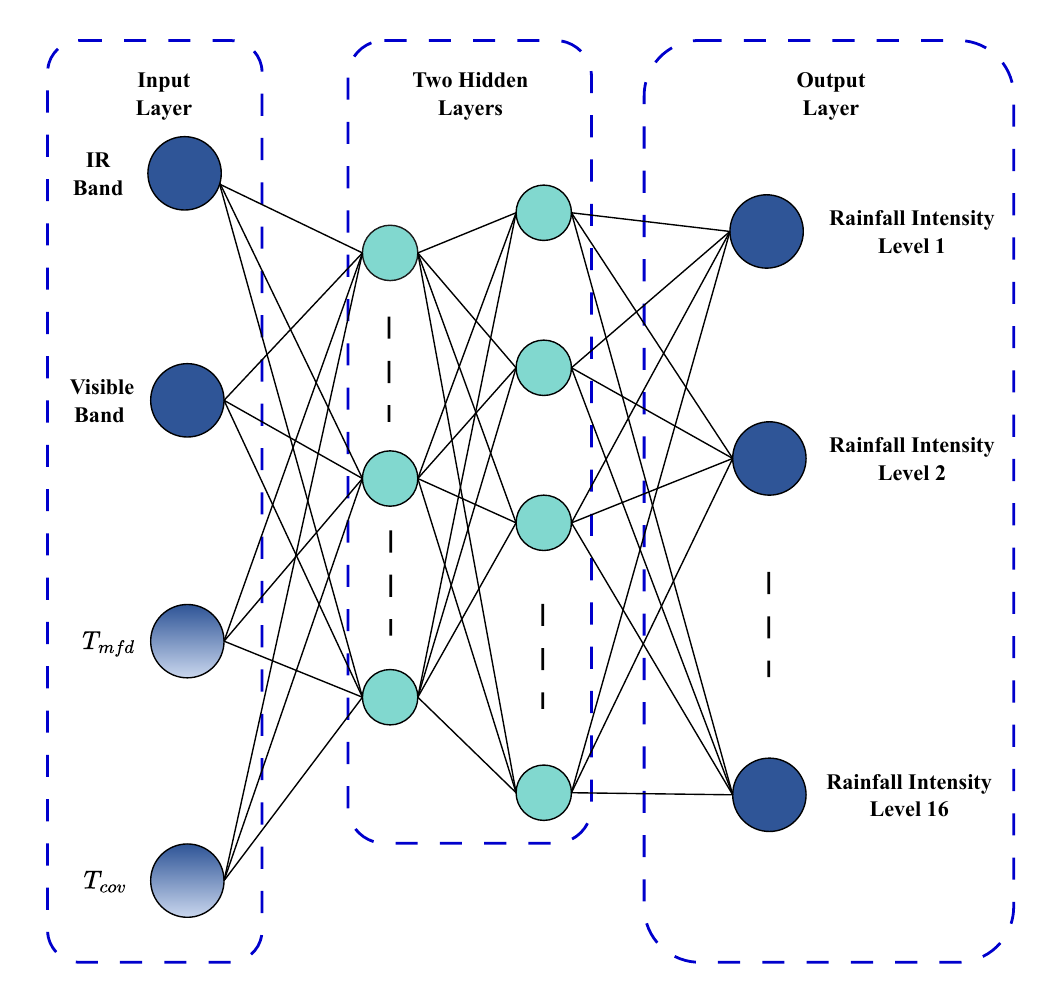}
\caption{Architecture of the ANN-based precipitation prediction model proposed by Chen et al.\cite{TaoChen1993}.}
\label{ANN Model}
\end{figure}

In 1993, Chen et al. proposed a 4-layer ANN model on rainfall prediction using geostationary meteorological satellite (GMS) image sets of Shikoku, Japan\cite{TaoChen1993}. They introduced $T_{cov}$ and $T_{mfd}$ to complement IR temperature and visible image graylevel as input features (shown in Fig.~\ref{ANN Model}).

In detail, $T_{cov}$ represents the division of variation and the average of the $2^\circ$ length region, while $T_{mfd}$ represents the average difference computed along two diagonal directions of the image grid, namely northwest–southeast (NW–SE) and northeast–southwest (NE–SW), within 3 grid distances. The proposed network has 2 hidden layers, possessing the ability to sort the relationships with the input data (e.g., IR band, visible band, and input features). During the training process, rainfall intensity of ground true radar was used as the supervised labels, with randomly selecting half of the original GMS images as a training set. By reducing the rainfall intensity to 4 levels, the model achieved a converged output with a good degree of intensity differentiation. They got a 90.45$\%$ accuracy score in validation data, comparing higher than regression based conventional methods.

In recent years, ANN has still played an important role in rainfall prediction, with some improved variant methods being used\cite{Junaida2012}\cite{Ogunrinde2020}. Ahuna et al. presented and validated a backpropagation artificial neural network for short-term rainfall rate prediction with a primary focus on mitigating rain fade over Earth-satellite communication links\cite{Ahuna2017}. The goal of their network is to predict a rainfall rate $x(t+1)$ estimated by the following expression,
\begin{equation*}
x(t+1)=f(x(t),x(t-1),x(t-2))
\end{equation*}
where $x(t)$ is the current rainfall rate, $x(t-1)$ and $x(t-2)$ are previous rainfall rates at time $(t-1)$ and $(t-2)$ respectively. Through experimental evaluation, a 3:5:1 network architecture (three inputs, five hidden neurons, and one output) achieved the best performance with an MSE of 1.1063 at epoch 30, demonstrating the feasibility of neural-network-based real-time rainfall prediction for satellite communication applications. 

DNNs are a subset of ANNs distinguished by their depth, typically defined as having multiple (often more than three) hidden layers. The additional layers allow DNNs to learn hierarchical feature representations. However, this architectural complexity requires precise configuration to be effective. For instance, Peng et al. enhanced neural network-based precipitation models by integrating a fuzzy control-based multicellular gene programming (FMCGEP) framework into a DNN architecture\cite{Peng2022}. This method systematically searches the parameter space to identify optimal architectures. The optimized model, FMCGEP-DNN, demonstrated higher accuracy and robustness, highlighting that automated optimization can mitigate human bias in model design and enhance generalization under different precipitation conditions. 

Although network depth implies superior representational power, practical results can vary depending on data constraints. Rayudu et al. demonstrated this in a comparison between ANN and DNN, finding that a well-optimized shallow ANN actually surpassed a DNN in predictive accuracy for certain rainfall datasets\cite{Rayudu2023}. This indicates that for tasks with limited data volume or specific regional characteristics, the hierarchical advantages of deep neural networks may be offset by the increased risk of overfitting and the increased computational cost. Consequently, the choice between shallow and deep architectures remains a balance between a model's representational capacity and the available training resources. Finally, the key characteristics of these models are summarized in Table~\ref{ANNDNN_comparison}.

\begin{table*}[t]
\caption{Brief comparison of ANN/DNN models in precipitation prediction}
\label{ANNDNN_comparison}
\centering
\small
\renewcommand{\arraystretch}{1.25}
\setlength\tabcolsep{0.9pt} 
\begin{tabular}{%
L{0.15\textwidth}
L{0.15\textwidth}
L{0.15\textwidth}
L{0.27\textwidth}
L{0.27\textwidth}}
\hline
\textbf{Nerual Network} &\textbf{Prediction Target} & \textbf{Data} & \textbf{Comparison vs baselines (bold)} & \textbf{Contribution} \\
\hline
Four-layer ANN\cite{TaoChen1993} & Rainfall intensity classes & GMS satellite images and radar data & Improvement over \textbf{regression methods} & One of the earliest works linking satellite imagery directly to rainfall intensity using ANN \\
\hline
Three-layer ANN with stepwise regression for input selection\cite{Junaida2012} & Extreme precipitation events & Malaysian meteorological data and climate indices & Input variable selection (IVS) improves \textbf{standard ANN} performance vs using all inputs & Introduced IVS before ANN training to improve heavy rainfall prediction\\
\hline
Three-layer ANN\cite{Ogunrinde2020} & Standardized precipitation and evapotranspiration index (SPEI) & Nigerian station data & Achieved 0.82 in $R^2$, 0.75 in RMSE, 0.79 in NSE, and 0.56 in MSE, outperforming \textbf{statistical methods} (0.51, 0.57, 0.28, 0.44, respectively) & Forecasted SPEI using ANN combining meteorology and climate indices\\
\hline
Three-layer ANN\cite{Ahuna2017} & 30-second ahead rainfall rate & Durban rainfall measurements & ANN prediction errors kept within acceptable RMSE & Provided relatively accurate short-term rainfall prediction\\
\hline
DNN with FMCGEP-based hyperparameter optimization\cite{Peng2022} & Daily precipitation statistics & 3 datasets with different characteristic attributes & Outperformed \textbf{MLR, BP, SVM, RF, and standard DNN} (e.g., improves $8.11\%, 3.78\%, 10.66\%, 3.04\%$ in MSE, RMSE, $R^2$ and MAE on the dataset 1) & Automatic hyperparameter optimization of DNN for precipitation prediction using evolutionary programming\\

\hline
\end{tabular}
\end{table*}

\subsubsection{CNN Based Models}
Although ANN models solved the non-linear regression problem in prediction, higher classification accuracy is expected when more image feature is used. Consequently, CNNs\cite{Lecun1998}, which possess the ability to effectively exploit spatial context by incorporating neighborhood pixel information, have attracted the attention of researchers in recent years. Generally, CNNs consist of multiple convolutional layers and pooling layers, and combine nonlinear activation functions and fully connected layers before the output. Convolutional layers and pooling layers can automatically and adaptively learn features through convolution and pooling operations. This design enables the network to capture low-level visual patterns in the early stage, gradually abstract these patterns through subsequent processing, and eventually integrate them into high-level representations suitable for classification or regression tasks.

\begin{figure}
\includegraphics[width=1\linewidth]{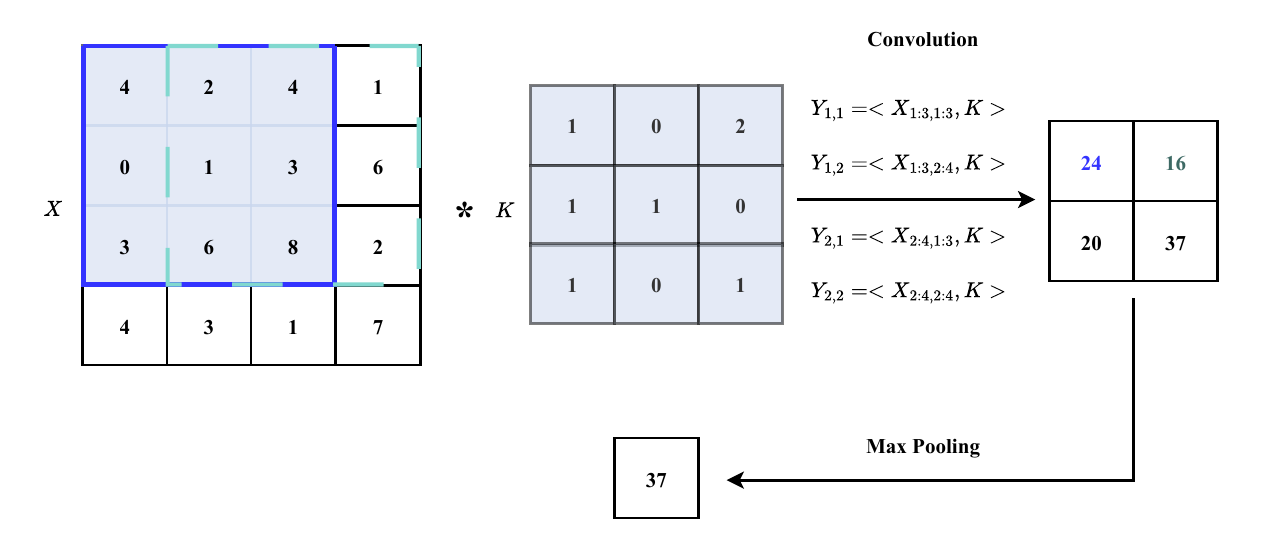}
\caption{Convolution and Max Pooling.}
\label{Convolution calculation process}
\end{figure}

As illustrated in Figure~\ref{Convolution calculation process}, the convolution operation in a CNN uses a simple 4 $\times$ 4 input image patch $X$ and a 3 $\times$ 3 kernel $K$. The kernel is sequentially applied to overlap subregions of the input through cross-correlation, where corresponding elements are multiplied and summed to produce a single scalar value. Specifically, the highlighted 3 $\times$ 3 region in the upper-left corner of the input is multiplied elementwise with the kernel, yielding the output value $Y_{1,1}=24$. By shifting the kernel with one stride, the process generates subsequent values such as $Y_{1,2}=16$, eventually producing the complete 2$ \times$ 2 feature map. After the convolution step, a pooling operation is applied to further downsample the feature map. Pooling reduces the spatial resolution while retaining the most important information, thereby improving computational efficiency and providing a certain degree of translational invariance. In this case, the pooled output is resized to a single scalar value $max\{24,16,20,37\}=37$. Alternatively, average pooling could be used to compute the mean value within each region. From the above steps, it can be seen that the pooling operation can gradually summarize the feature representations learned by the network.

A typical example is a purely CNN-based model using radar composite data from DWD to predict rainfall in Germany\cite{Ayzel2020}. The CNN architecture captures spatial correlations in consecutive radar frames, taking recent radar reflectivity maps as input and directly predicts precipitation fields for future lead times. The evaluation showed that the CNN produced more accurate precipitation predictions than both persistence and optical flow, especially for convective rainfall.
Kim et al. investigated whether CNNs can extend NASA’s IMERG Early Run satellite rainfall product (normally released with a 4-hour latency) into a near–real-time hurricane rainfall forecasting tool\cite{KIM2022}. Using a dataset of 37 Gulf-region hurricane events (2002–2019), the authors developed and compared two CNN architectures, namely a basic CNN model (no pooling) and a standard CNN model (with pooling and transposed convolution). The result shows both models accurately captured the spatial extent and structural evolution of hurricane rainfall fields, and basic CNN consistently produces slightly better accuracy and lower errors, demonstrating that even a smaller and simpler CNN could show better performance. In addition to short-term forecasting ability, CNNs also performs well in long-term predictions. Barnes et al. developed a CNN model which uses forecasted mean sea-level pressure (MSLP) and 2-m air temperature (2AT) anomaly fields from ECMWF as inputs, and CEH-GEAR (Centre for Ecology and Hydrology, gridded estimates of areal rainfall\cite{Tanguy2019}) regional rainfall as ground truth\cite{Barnes2023}. They tested the model on monthly and regional rainfall forecasts for Great Britain at seasonal to sub-seasonal lead times (1, 3, and 6 months). They compared the developed CNN with ECMWF predictions, and the result shows CNN outperformed the latter across all lead times. Byun et al. employed CNNs to quantitatively predict rainfall intensity directly from cloud images captured by low-cost IoT sensors \cite{Byun2023}. By combining a binary classification for rainfall existence with  CNN-based image-value model, the authors demonstrated that cloud imagery can serve as a viable data source for short-timescale rainfall prediction. In addition, CNN also has enhanced quantitative precipitation estimation, demonstrating improved accuracy over traditional radar-based methods, particularly in regions with complex terrain\cite{ChengYung2023}. Here, we conclude exemplary CNN models in Table~\ref{CNN_comparison}.

\begin{table*}[t]
\caption{Brief comparison of CNN models in precipitation prediction}
\label{CNN_comparison}
\centering
\small
\renewcommand{\arraystretch}{1.25}
\setlength\tabcolsep{0.9pt} 
\begin{tabular}{%
L{0.15\textwidth}
L{0.15\textwidth}
L{0.15\textwidth}
L{0.27\textwidth}
L{0.27\textwidth}}
\hline
\textbf{Nerual Network} &\textbf{Prediction Target} & \textbf{Data} & \textbf{Comparison vs baselines (bold)} & \textbf{Contribution} \\
\hline
RainNet (encoder–decoder CNN with skip connections)\cite{Ayzel2020} & 5-60 min lead time rainfall forecast & German DWD radar composites & Outperforms \textbf{the Eulerian persistence model and the Dense model} in MAE and CSI & Introduced CNN-based radar short-term forecasting replacing optical-flow extrapolation\\
\hline
CNN models with/without pooling layers \cite{KIM2022} & 30 min lead time hurricane rainfall forecasting & Rainfall imagery from NASA's IMERG-E product &  CNN model without pooling layers performs slightly better than \textbf{the model with pooling layers} & Proved CNN models available for short-term  hurricane rainfall forecasting fields \\
\hline
CNN mapping atmospheric predictors to rainfall totals \cite{Barnes2023} & Monthly rainfall forecasting & ECMWF SEAS5 and CEH-GEAR & The CNN achieves lower RMSE in 1, 3, 6-month lead times than \textbf{the ECMWF SEAS5 model} & First demonstration that CNNs can enhance monthly regional rainfall forecasts by learning spatial atmospheric circulation patterns \\
\hline
Binary classification model with CNN-based image-value model \cite{Byun2023} & Rainfall intensity prediction & Camera images & 85.59\% of the predicted results are within the 1 mm/h range from the observed values (no baseline) & First CNN “image-value” model predicting rainfall directly from ground cloud images\\
\hline
\end{tabular}
\end{table*}

\subsubsection{RNN and LSTM Based Models}
\begin{figure*}[!t]
\centering
\includegraphics[width=1\linewidth]{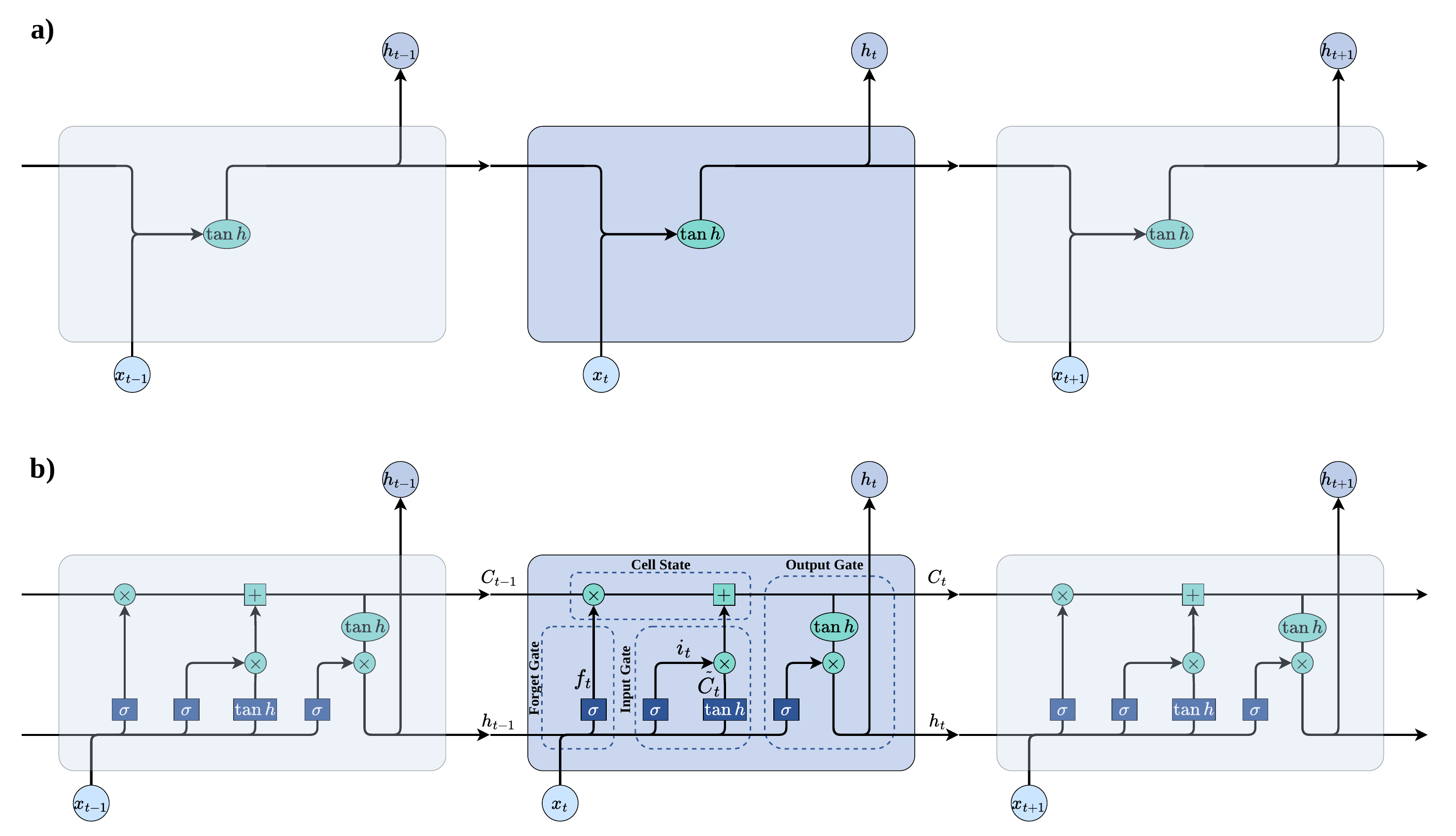}
\caption{RNN and LSTM model\cite{Fang2021}. a) Standard RNN structure models sequential evolution through recursive hidden state updates.  b) LSTM model incorporates gating mechanisms (input, forget, output gates) and a cell state to preserve long-term temporal dependencies.}
\hfill
\label{RNN and LSTM}
\end{figure*}
As shown in Figure~\ref{RNN and LSTM}, a RNN is designed to process sequential data by maintaining a hidden state that captures information from previous time steps. Unlike feedforward networks, RNNs incorporate recurrent connections, enabling the transfer of information forward in time, which makes them well-suited for tasks where the order of inputs is important, such as time series forecasting, speech recognition, natural language processing, and meteorological prediction\cite{ELMAN1990}, \cite{Goodfellow2016}.

Mathematically, at each time step $t$, the hidden state $h_t$ is updated according to
\[
h_t = f(W_x x_t + W_h h_{t-1} + b),
\]
where $W_x$ and $W_h$ are weight matrices, $b$ is a bias term, and $f$ is a nonlinear activation function (e.g., $\tanh$ or ReLU).

Relying on the power of RNN, some innovative approaches have been introduced to precipitation prediction. Ma et al. proposed Multimodal RNN (MM-RNN) , which integrates multiple meteorological data streams such as radar reflectivity, ground weather station data, and atmospheric reanalysis variables into its prediction framework\cite{Ma2023}. This model consists of three modules: the meteorological elements’ prediction module (EM), the radar prediction module (RM), and the modal fusion module (MFM). EM and RM apply RNN structures to capture their unique temporal dynamics, while MFM combines the hidden states from all modalities before the final output layer, allowing the network to learn complementary relationships between atmospheric variables. By training and evaluating with MeteoNet and RAIN-F datasets, the multimodal architecture shows better performance across most metrics compared with other unimodal RNN models. 

While RNNs are capable of modeling temporal dependencies, standard RNNs encounter difficulties with long-term dependencies due to the vanishing or exploding gradient problem. This limitation has motivated the development of LSTMs\cite{Hochreiter1997}, which use gating mechanisms to selectively retain or discard information, thereby enabling more efficient modeling of longer-range dependencies.
A standard RNN only have a simple repeating component, such as a tanh layer. In contrast, the corresponding component in LSTM has 4 neural network layers, namely, 3 sigmoid layers and a tanh layer. 
In detail, the unique structure in LSTM is called gate\cite{Gers1999}, which consists of a sigmoid layer and a pointwise operation. The first gate, called the “forget gate,” decides what to throw away from the old cell state: 
\[
f_t = \sigma(w_f\cdot[h_{t-1},x_t]+b_f),
\]
where $\sigma$ represents a sigmoid function that controls the proportion of remaining information according to $h_{t-1}$ and $x_t$. $0$ means totally ignore, while $1$ means keeping.

Then, choose information from the current cell state for storing:
\[
i_t = \sigma(w_i\cdot[h_{t-1},x_t]+b_i),
\]
\[
\label{store_equ2}
\tilde{C_t} = \tanh(w_c\cdot[h_{t-1},x_t]+b_c),
\]
where $i_t$ decides what values need to be updated according to the current input, followed by a vector of candidate values $\tilde{C_t}$.

The next step is to combine them. Multiply the values inherited from the older state $C_{t-1}$, and the things that we want to forget earlier, then add $i_t*\tilde{C_t}$, where $*$ denotes pointwise operation (e.g., Hadamard product):
\[
{C_t} = f_t*C_{t-1}+i_t*\tilde{C_t}.
\]
Finally, the output $h_t$ is decided by a filtered operation of the current state and the result of a sigmoid layer:
\[
{o_t} = \sigma(w_o\cdot[h_{t-1},x_t]+b_o),
\]
\[
h_t=o_t*\tanh(C_t).
\]

To explore potential applicabilities of LSTM, Pathan et al. proposed an efficient LSTM-based framework for forecasting precipitation using historical rainfall observations\cite{Pathan2021}. The model utilized monthly precipitation data from 1990 to 2020 from the NOAA Climate Data Online service to generate 1-day-ahead forecasts. To enhance data quality and stabilize variance, a Box-Cox transformation was applied to the time series, effectively reducing skewness to facilitate more accurate patterns for training. The designed network, comprising two recurrent layers and approximately 13,000 trainable parameters, achieved a competitive RMSE of 6.181, outperforming traditional persistence and average forecast benchmarks while significantly reducing computational cost. 
Wang et al. applied LSTM to predict local precipitation in Guizhou by using multi-source data from Doppler weather radar, laser raindrop spectrometers, and automatic rain gauges\cite{Wang2023}. Although the method based on LSTM exhibited a slightly larger cumulative error and has a significant lag at the beginning of precipitation, it showed a higher overall correlation coefficient (CC) compared to the traditional dynamic $Z-I$ relationship. To better comprehend these models, key characteristics are summarized in Table~\ref{RNNLSTM_comparison}.

\begin{table*}[t]
\caption{Brief comparison of exemplary RNN/LSTM models in precipitation prediction}
\label{RNNLSTM_comparison}
\centering
\small
\renewcommand{\arraystretch}{1.25}
\setlength\tabcolsep{0.9pt} 
\begin{tabular}{%
L{0.15\textwidth}
L{0.15\textwidth}
L{0.15\textwidth}
L{0.27\textwidth}
L{0.27\textwidth}}
\hline
\textbf{Nerual Network} &\textbf{Prediction Target} & \textbf{Data} & \textbf{Comparison vs baselines (bold)} & \textbf{Contribution} \\
\hline
Two-layer LSTM \cite{Pathan2021} & 1-day ahead forecasting &  Monthly precipitation data & RMSE (6.181) is lower than \textbf{persistence} (7.967) and \textbf{average} (6.192) forecasting methods & Provided a lightweight framework for efficient univariate precipitation forecasting\\
\hline
Multimodal RNN \cite{Ma2023} & Short-term (few hours ahead) forecasting & MeteoNet and RAIN-F radar datasets & Superior to \textbf{multiscale RNN} in both datasets & Introduced multimodal structure for fusion of radar and atmospheric variables\\
\hline
\end{tabular}
\end{table*}

\subsubsection{GNN Based Models}
To capture and learn from the irregular topology and relational dependencies inherent in graph-structured data, a novel class of model named GNN has been developed. Basically, a graph $G$ consists of a set of nodes $V$ and edges $E$, where each node can represent an entity and each edge represents a relationship or interaction between entities. The target of GNN is to learn a state embedding $h_v \in \mathbb{R}^s$ for each node, which includes features in node itself and relationship of its neighborhood, defined as: 
\[
\mathbf{h}_v=f(\mathbf{x}_v,\mathbf{x}_{e[v]},\mathbf{h}_{\mathcal{N}v}, \mathbf{x}_{\mathcal{N}v}),
\]
where $\mathbf{x}_v$, $\mathbf{x}_{e[v]}$, $\mathbf{h}_{\mathcal{N}v}$, $\mathbf{x}_{\mathcal{N}v}$ are the features of $v$, the features of its edges, the states and the features of the neighbored nodes, respectively. $f$ here is a parametric function named the local transition function. An output $\mathbf{o}_v$ could be produced by $\mathbf{h}_v$ and features $\mathbf{x}_v$ by a local output function $g$, denoted as:
\[
\mathbf{o}_v=g(\mathbf{h}_v,\mathbf{x}_v),
\]
In general, $f$ and $g$ can be interpreted as feedforward neural networks. All features and all states can be stacked in matrices $\mathbf{X}$ and $\mathbf{H}$, respectively, and a global transition function $F$ stacks all versions of $f$ for all nodes\cite{ZHOU2020}. To train $\mathbf{H}$, an iterative scheme denoted as:
\[
\mathbf{H}^{t+1}=F(\mathbf{H}^t,\mathbf{X}),
\]
where t is the -th iteration step. 

In recent years, researchers have taken GNN as the base and enhanced the model's understanding ability of physical information or radar data by adding some sub-modules. These sub-modules only serve as auxiliary functions, so we still consider them as GNN-based models. For example, Chen et al. proposed a physics-informed GNN for 6-hour accumulated precipitation forecasting over China, using the ERA5 reanalysis dataset and the merged precipitation observations\cite{Chen2024}. In their model, each graph node corresponded to a geographical grid cell containing multiple atmospheric predictors, including vertical velocity, relative humidity, geopotential height, and temperature fields, while edges encoded spatiotemporal correlations. To ensure physical consistency, they embedded large-scale circulation indices directly into the graph structure. Evaluation results at 12, 18, and 30-hour lead times demonstrated that this approach significantly outperforms traditional NWP models, as well as CNN and Transformer baselines. Notably, the model excels in capturing heavy rainfall events and maintaining regional precipitation variability. 

The Precipitation Nowcasting Network via Hypergraph Neural Networks (PN-HGNN), a novel precipitation nowcasting framework that combines a spatiotemporal attention mechanism with hypergraph convolution layers, was proposed to process spatial dependencies and temporal progression across multiple radar frames\cite{Sun2024}. The model was specifically designed for short-term nowcasting within a 2-hour window, and was evaluated using the Hong Kong Observatory (HKO)-7 dataset. Experimental results demonstrate that this hypergraph-based approach achieves superior prediction performance compared to 6 representative echo extrapolation models, addressing the common issue of blurred or implausible forecasts in traditional deep learning methods. Peng et al. introduced a multilevel rainfall forecasting framework based on structured GNN\cite{Peng2023}. This framework was separated into multiple submodels, which were trained for different rainfall thresholds (e.g., 0.1~mm, 3~mm, 10~mm, 20~mm). Each submodel utilizes the GNN architecture to propagate information spatially and temporally. Across all rainfall levels, the proposed model consistently outperformed ECMWF’s IFS in the threat score (TS) and the Heidke skill score (HSS), particularly at higher intensities. The result shows that the framework can serve as a post-processing enhancement for operational forecasts rather than a replacement of NWP, offering a computationally efficient route to improve rainfall prediction in real-world applications. Inspired by this research, Yousaf et al. designed a hybrid GNN-based framework to post-process and enhance local-scale numerical weather forecasts in the Calabria region of Italy\cite{Yousaf2025}. The proposed spatiotemporal GNN with six graph convolution layers and ReLU activations was trained using MSE loss, and it provided tangible improvements over classical Weather Research and Forecasting (WRF) outputs and other machine learning post-processors. Pham et al. employed the encoder-decoder paradigm, specifically, they used GNNs as an encoder and CNNs as a decoder, which interprets high-dimensional spatial-temporal information into a low-dimensional format and extracts features to produce the final precipitation estimation, respectively\cite{Pham2024}. Another famous GNN-based model, namely GraphCast, was developed by Google DeepMind Research, designed with an encoder-processor-decoder architecture\cite{Lam2023}. The encoder, processor, and decoder formed a coherent graph-based pipeline in which the encoder projected atmospheric variables onto a high-resolution multimesh representation, the processor performed hierarchical message passing to capture global spatial dependencies, and the decoder reconstructed the updated atmospheric state on the latitude--longitude grid as a normalized residual correction to the previous forecast step. Although the model has 36.7 million parameters, it achieved superior performance over the most accurate operational deterministic systems (ECMWF’s HRES) on approximately 90$\%$ of 1380 verification targets, demonstrating improved forecast skill across multiple atmospheric variables and lead times up to 10 days. To evaluate the ability of GraphCast’s precipitation prediction, Yan et al. validated the model against 2393 ground observation stations in mainland China\cite{Yan2025}. The result shows that GraphCast maintained forecast stability across increasing lead times, compared with ECMWF's forecasts in multiple metrics (e.g., CC, ME, RMSE, POD).
These studies highlight how the latest generation of GNNs overcomes the limitations of earlier architectures by effectively blending physical drivers with multi-source meteorological data. By linking graph nodes to atmospheric variables, these networks make precipitation forecasting more accurate and adaptable. We also briefly conclude exemplary GNN models in Table~\ref{GNN_comparison}.

\begin{table*}[t]
\caption{Brief comparison of GNN models in precipitation prediction}
\label{GNN_comparison}
\centering
\small
\renewcommand{\arraystretch}{1.25}
\setlength\tabcolsep{0.9pt} 
\begin{tabular}{%
L{0.15\textwidth}
L{0.15\textwidth}
L{0.15\textwidth}
L{0.27\textwidth}
L{0.27\textwidth}}
\hline
\textbf{Nerual Network} &\textbf{Prediction Target} & \textbf{Data} & \textbf{Comparison vs baselines (bold)} & \textbf{Contribution} \\
\hline
Physics-informed GNN \cite{Chen2024} & 6-hour accumulated precipitation of 12, 18, 30 hour ahead & ERA5 and merged precipitation observations & Outperforms \textbf{NWP, U-Net, CNN-3D}, especially for heavy rainfall prediction & Introduced a physics-informed GNN coupling meteorological variables to improve precipitation forecasts\\
\hline
PN-HGNN \cite{Sun2024} & Prediction within 2 hours & HKO-7 radar dataset & Better than \textbf{Optical flow, ConvLSTM, MotionRNN, PredRNN, MIM, REMNet} in POD & Demonstrated hypergraph neural networks to capture high-order spatial correlations in radar echo evolution\\
\hline
\end{tabular}
\end{table*}

\subsubsection{GAN Based Models}
\begin{figure}
    \centering
    \includegraphics[width=1\linewidth]{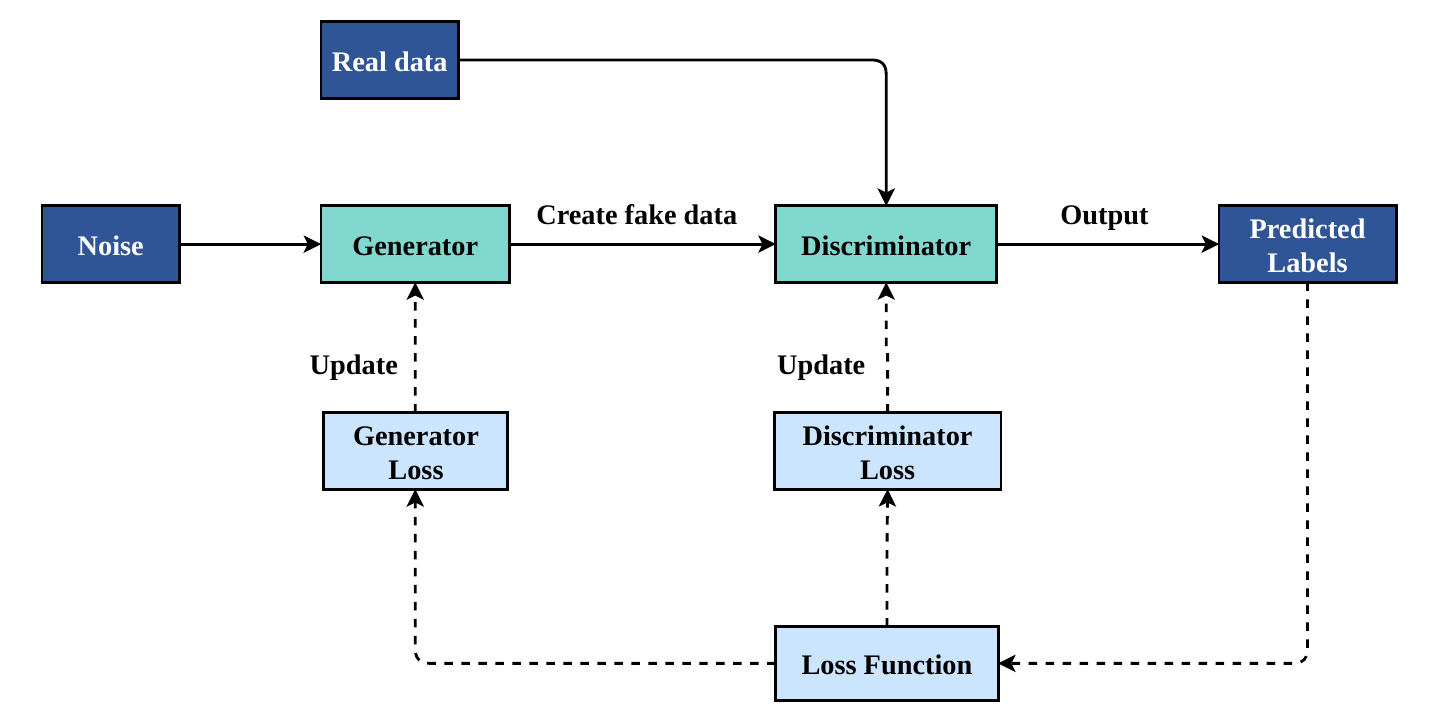}
    \caption{Schematic of the adversarial training strategy in GAN-based precipitation prediction. The generator produces synthetic fields from random noise, while the discriminator evaluates their realism against observed data. The adversarial loss drives iterative updates of both networks\cite{Pachika2024}.}
    \label{fig:GAN}
\end{figure}

Generative Adversarial Networks (GANs), introduced by Goodfellow et al.\cite{Goodfellow2014}, are a class of generative models that learn to synthesize realistic data through an adversarial training process between two neural networks: a Generator $G$ and a Discriminator $D$. As shown in Figure~\ref{fig:GAN}, the generator aims to produce data samples that mimic the real data distribution, while the discriminator attempts to distinguish between real and generated samples. This interaction can be formulated as a minimax optimization problem:
\[
\min_G  \max_D \; \mathbb{E}_{x \sim p_{\text{data}}(x)} \left[ \log D(x) \right] 
+ \mathbb{E}_{z \sim p_z(z)} \left[ \log \left( 1 - D(G(z)) \right) \right],
\]
where $p_{data}(x)$ is the real data distribution and $p_z(x)$ is the prior distribution of the generator’s input noise vector $z$. So, in an adversarial training loop, $G$ and $D$ are trained alternately, $G$ improves to fool $D$, and $D$ improves to detect fake data.

In precipitation forecasting, GANs are often implemented in a conditional form (cGANs), where the generator is conditioned on input meteorological fields to produce future precipitation maps. It’s an extension of the original GAN, but instead of generating samples purely from random noise, the generation process is conditioned on some extra information. The formulation is denoted as:
\begin{align*} 
& \min_G \max_D \; \mathbb{E}_{x \sim p_{\text{data}}(x)} \left[ \log D(x|y) \right] \\
&+ \mathbb{E}_{z \sim p_z(z)} \left[ \log \left( 1 - D(G(z|y)) \right) \right],
\end{align*}
where $y$ is a condition. As an exemplary application, Price and Rasp introduced CorrectorGAN, a conditional GAN designed to simultaneously correct biases and super-resolve coarse global ensemble forecasts, producing high-resolution probabilistic precipitation fields that approach the skill of operational regional models while significantly reducing computational cost\cite{price22a}. To further address mode collapse and heterogeneous precipitation patterns, An et al. developed a self-clustered GAN framework (ClusterCast) that introduces self-supervised clustering within the generator to learn disparate latent representations of precipitation types, improving robustness across convective and stratiform regimes\cite{An2024}. Furthermore, PredAt-GAN employs adversarial training to encourage sharper radar echo reconstruction while preserving fine-scale structures of extreme precipitation systems. The discriminator guides the generator to produce spatially coherent and temporally consistent forecasts, thereby improving both visual fidelity and quantitative precipitation skill scores\cite{Ji2024}\cite{KimGAN2022}.
To better highlight the practical challenges addressed by GANs, Table~\ref{GAN_comparison} summarizes the core problem focus, prediction target, datasets, baseline comparisons, and key contributions of representative models.

\begin{table*}[t]
\caption{Brief comparison of GAN models in precipitation prediction}
\label{GAN_comparison}
\centering
\small
\renewcommand{\arraystretch}{1.25}
\setlength\tabcolsep{0.9pt} 
\begin{tabular}{%
L{0.14\textwidth}
L{0.14\textwidth}
L{0.15\textwidth}
L{0.15\textwidth}
L{0.195\textwidth}
L{0.2\textwidth}}
\hline
\textbf{Nerual Network} &\textbf{Prediction Target} & \textbf{Data} & \textbf{Benchmarks} & \textbf{Best Score} & \textbf{Contribution} \\
\hline
ClusterCast\cite{An2024} & Next 2 hours & Radar frames & Rainymotion, ConvLSTM, TrajGRU, DGMR & 0.383(CSI), 0.614(FSS), 0.329(ETS), 0.476(HSS) on 120 minutes prediction with 1 mm/h threshold  & Proposed a self‑clustered generator and two discriminators (spatial and temporal) to mitigate mode collapse in non-stationary rainfall\\
\hline
PredAT-GAN \cite{Ji2024} & Next 60 minutes & Radar echo dataset & T-Unet-TrajGRU, RainPredRNN, PFST-LSTM, SmaAT-Unet, PredRNN\_v2 &  0.4113 (average CSI), 0.4538 (average HSS) & Improved heavy rainfall prediction via attention-guided adversarial learning\\
\hline
CorrectorGAN \cite{price22a} & Accumulated precipitation in 6-12 hours & TIGGE/weather-radar estimates & HREF, TIGGE, Pure-SR GAN, BG-CNN & Close to HREF (e.g., 0.574 vs 0.562 in CRPS)& First GAN to jointly perform bias correction + super-resolution for global precipitation forecasts \\
\hline
\end{tabular}
\end{table*}

\subsubsection{Transformer Based Models}
RNNs and LSTMs have historically been the backbone of sequence modeling tasks due to their ability to capture temporal dependencies through sequential processing. However, as the significant growth in data size of input, these models suffer from inherent limitations, including difficulty modeling long-range dependencies, limited parallelization during training, and gradient vanishing or exploding issues. A neural network architecture based on self-attention mechanisms was introduced to enable large-scale parallelism and scalability, namely Transformer\cite{Vaswani2017}. 
The Transformer architecture is a deep learning model designed to process sequential data without relying on recurrence. Instead, it leverages a self-attention mechanism to model dependencies between sequence elements, allowing all tokens to be processed in parallel\cite{Aleissaee2023}. This innovation greatly improves training efficiency and the ability to capture long-range dependencies compared to traditional recurrent neural networks.
\begin{figure}
    \centering
    \includegraphics[width=1\linewidth]{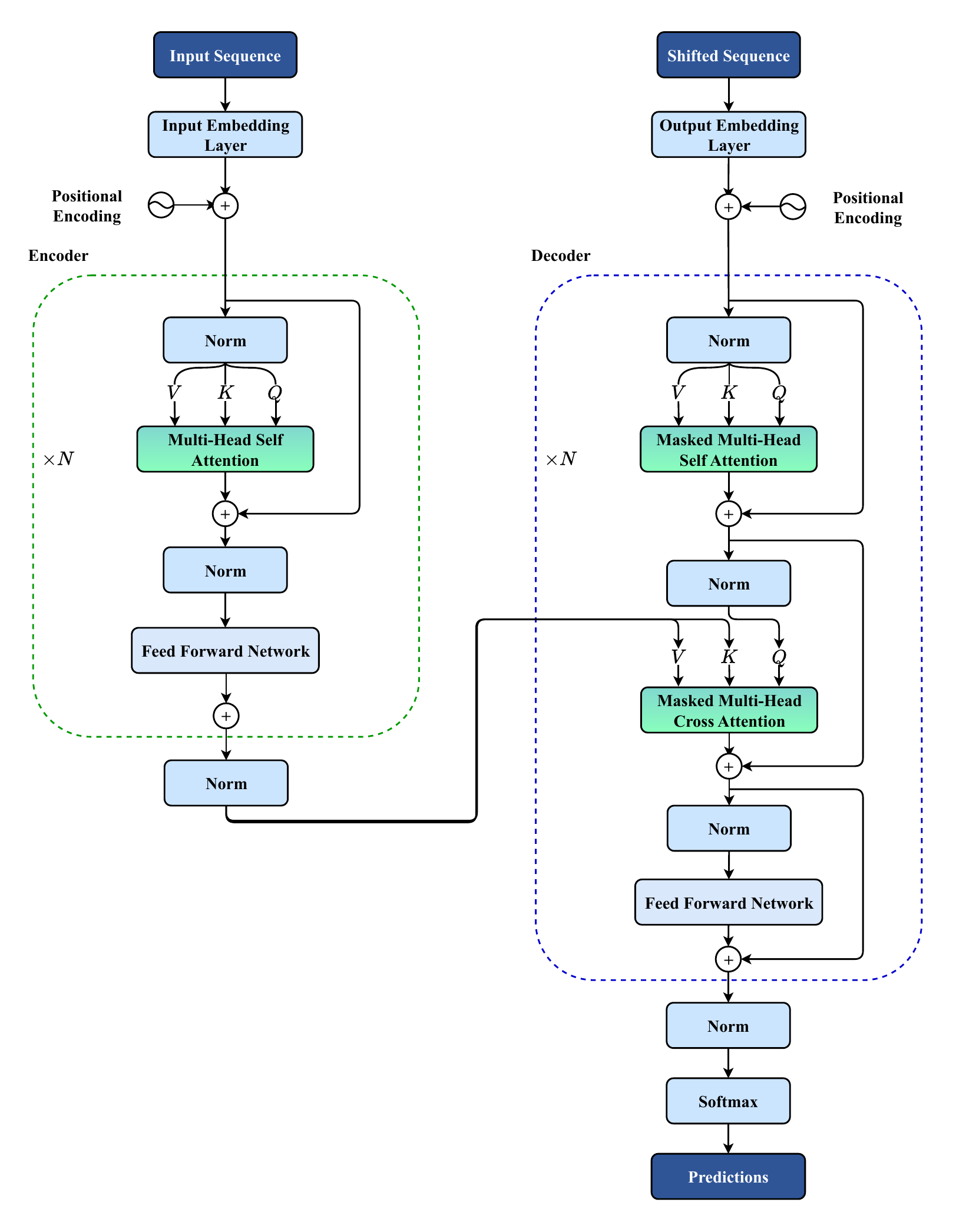}
    \caption{Transformer Architecture\cite{Vaswani2017}, \cite{Aleissaee2023}. The encoder processes the input sequence through stacked multi-head self-attention and feed-forward layers with positional encoding, enabling global dependency modeling. The decoder incorporates masked self-attention and cross-attention to generate future predictions.}
    \label{fig:Transformer}
\end{figure}
Refer to Figure~\ref{fig:Transformer}, a Transformer consists of two primary components: the encoder and the decoder. The encoder transforms the input sequence into a contextual representation through stacked layers of multi-head self-attention and position-wise feed-forward networks, each followed by residual connections and layer normalization. The decoder generates the output sequence autoregressively, using masked multi-head self-attention to prevent access to future positions, cross-attention to incorporate encoder outputs, and feed-forward layers for transformation. The core of Transformer is the attention mechanism, which is a method for dynamically focusing on the most relevant parts of the input when producing each output.
Mathematically, in standard attention:
\[
Attention(Q, K, V)=softmax({\frac{QK^T}{\sqrt{d_k}}})V,
\]
where Q represents for queries, K represents for keys, V represents for values, $d_k$ is the dimensionality of the key vectors. For spatiotemporal meteorological data, precipitation at a given grid cell may depend heavily on patterns far away (e.g., an incoming storm). Attention enables the model to dynamically capture both local and long-range dependencies without fixed kernel sizes or recurrence.

Based on the power of Transformer, a novel spatiotemporal SwinUNet3D model was proposed for short-term weather forecasting, leveraging the 3D Swin-transformer in a U-net architecture to predict 8 hours ahead weather events from hourly gridded data\cite{Bojesomo2021}. By replacing all convolutional layers in the traditional U-Net with 3D Swin Transformer blocks, the model outperforms traditional U-Net and persistence baselines in the core and transfer tasks of the IEEE Big Data Weather4cast Competition. As Transformer unfolded competitive accuracy and computational efficiency, more and more projects harnessed it to address weather forecasting tasks. Earthformer, a novel space-time Transformer model designed for Earth system forecasting\cite{Gao2022}. The model is built on a custom Cuboid Attention mechanism that dramatically reduces the computational cost of self-attention in high-dimensional spatiotemporal data. Earthformer achieves state-of-the-art performance across real-world precipitation nowcasting (SEVIR dataset) and long-term sea surface temperature forecasting (ICAR-ENSO dataset). Another model, named Preformer, is a lightweight transformer-based framework specifically designed for precipitation nowcasting, adopts a minimalist encoder-translator-decoder architecture with only two Transformer layers, outperforms both traditional and modern baselines (e.g., ConvLSTM, SimVP, TAU, gSTA) on ERA5 and WeatherBench datasets\cite{Jin2024}. Similarly, the recently proposed SwinNowcast model also demonstrates the potential of transformer-based architectures for precipitation nowcasting\cite{LiZhuang2025}. Built upon the Swin Transformer’s hierarchical representation learning and shifted window attention, SwinNowcast captures both local and global spatiotemporal dependencies in radar echo sequences. Extensive evaluations on benchmark datasets indicate that it surpasses leading baselines, including ConvLSTM and PredRNN, in predictive accuracy and the visual realism of forecasted precipitation maps. Multi-Scale Spatial-Temporal Transformer (MSTT)\cite{LiTian2025}, a model for meteorological variable forecasting, especially well-suited for precipitation prediction, addressing the challenge of capturing both global and local dependencies in spatiotemporal climate data. 

Additionally, Li et al. expanded Transformer to Quantitative Precipitation Nowcasting (QPN)\cite{Li2023}, which aims to predict short-term, high-resolution precipitation fields based on recent radar echo observations, formulated as below.
\begin{align*} 
&\hat {X}_{t+1},\ldots,\hat {X}_{t+m} = \\ &\mathop {argmax}_{X_{t+1},\ldots,X_{t+m}} \quad p\left ({X_{t+1},\ldots,X_{t+m}|X_{t-n},X_{t-3},\ldots,X_{t} }\right), 
\end{align*}
where ${X_i}$ is the actual frame at time ${i}$, and $\hat {X}_j$ is the predicted frame at time $j$. $m$ and $n$ are the amount of input frames and predicted frames respectively. The proposed LPT-QPN architecture leverages a Transformer backbone augmented with physics-informed attention, where the standard self-attention mechanism is constrained by the advection equation, a fundamental physical law describing the motion of precipitation fields. Specifically, what is obviously different from previous researchers, the authors introduced a new attention mechanism named Multihead Squared Attention (MHSA) in order to reduce the computational complexity of the QPN task.
\begin{align*} &~\hat {\mathbf {X}}=W_{1}\cdot \text { reshape}\left ({\mathrm {Attention}\left ({\hat {\mathbf {Q}}, \hat {\mathbf {K}}, \hat {\mathbf {V}}}\right)}\right)+\mathbf {X},& \\ &~\mathrm {Attention}\left ({\hat {\mathbf {Q}}, \hat {\mathbf {K}}, \hat {\mathbf {V}}}\right)= \mathrm {Sigmoid}\left ({\hat {\mathbf {Q}} \cdot \hat {\mathbf {K}} ^{T}/ \alpha }\right)\cdot \hat {\mathbf {V}},& \\ &~\hat {\mathbf {Q}}=\text {reshape}\left ({W_{1}^{Q}W_{3}^{Q}\mathbf {X}_{\text {LN}}}\right), & \\ &~\hat {\mathbf {K}}=\text {reshape}\left ({W_{1}^{K}W_{3}^{K}\mathbf {X}_{\text {LN}}}\right),& \\ &~\hat {\mathbf {V}}=\text {reshape}\left ({W_{1}^{V}W_{3}^{V}\mathbf {X}_{\text {LN}}}\right),& \\ &~\hat {\mathbf {Q}},\hat {\mathbf {K}},\hat {\mathbf {V}}\in \mathbb {R}^{C\times H\cdot W},\mathbf {X}\in \mathbb {R}^{C\times H\times W}, \end{align*}
where, $\hat{Q}$, $\hat{K}$, $\hat{V}$ are the query, key and value matrices reshaped from feature maps of previous 1$\times$1 point-wise convolutions and 3$\times$3 depth-wise convolutions, i.e. $W_1^{(\cdot)}$, $W_3^{(\cdot)}$. $X_{LN}$ is the layer normalized input tensor and $\alpha$ is the trainable parameter. Here, a sigmoid activation function instead of softmax was applied to make the computation process more suitable for feature maps of precipitation data, because the sum of attention weights is not necessarily equal to 1. Comparative experiments on benchmark radar datasets demonstrate that LPT-QPN achieves state-of-the-art or near state-of-the-art performance while using fewer parameters and computational resources than deep recurrent models (e.g., ConvLSTM, PredRNN) and purely data-driven Transformers (e.g., Swin Transformer, SimVP). The model particularly excels in preserving fine-scale precipitation structures and reducing displacement errors over longer lead times.

As an emerging structure in machine learning, there are more and more transformer-based models applied in precipitation prediction in recent years. For example, Piran et al. applied a transformer generative framework for precipitation nowcasting to the Soyang Dam basin in South Korea, demonstrating superior performance compared to default models (cGANs, ConvLSTMs, U-Net and pySTEP), as validated through multiple verification metrics including PCC, RMSE, NSE, CSI and FSS\cite{PIRAN2024}. Zhao et al. presented a spatiotemporal transformer-based denoising diffusion model designed to advance the realism and physical consistency of precipitation nowcasting\cite{Zhao2024}. They experimented with the SEVIR dataset and finally found that the proposed approach yielded visually sharper, more realistic forecasts and achieved competitive meteorological evaluation scores across a range of rainfall intensities. Li et al. introduced a diffusion transformer framework with causal attention for high-resolution precipitation nowcasting, taking advantage of causal attention to ensure strict temporal ordering and directional information flow\cite{Li2024}. UTrans-Net, a U-shaped neural architecture integrating transformer attention mechanisms, was proposed by Cao et al. in 2022\cite{Cao2022}. The neural architecture was evaluated on the China Meteorological Administration dataset, and the result highlighted the model’s strong generalization ability and suitability for real-time forecasting applications. MDTNet, namely multiscale deformable transformer network, uses Deformable Multi-Head Self-Attention (D-MHSA), allowing spatially adaptive attention to strong precipitation regions\cite{Zhaozewei2024}. Particularly, the researchers novelly introduced Fourier space regularization and adversarial losses that constrain the model to reproduce high-frequency spectral components. Dong et al. presented a transformer-based framework designed to enhance radar echo extrapolation by unifying global--local spatiotemporal aggregation with motion-aware learning\cite{Dong2022}. HTLA, Hierarchical Transformer With Lightweight Attention, was proposed to efficiently nowcast precipitation based on the KNMI (Koninklijk Nederlands Meteorologisch Instituut) radar dataset, highlighted by Cross-Channel Self-Attention (CSA), Dual Feedforward Network (DFN) and Gaussian Pooling Skip-Connection strategy\cite{LiWenhui2024}. PFformer, a transformer-based model tailored for short-term precipitation forecasting in northern Xinjiang, China, introduces a piecewise-attention Transformer architecture for efficient short-term rainfall prediction\cite{Xu2024}. The core innovation of this study lies in replacing the point-wise dot product self-attention of iTransformer with a Segment Correlation Attention (SCAttention) mechanism. This modification reduces computational complexity from $O(L^2)$ to $O(L^2/L_{seg})$ ($L$ and $L_{seg}$ are the length of the time series and the divided length by multiple segments, respectively) by computing correlations between temporal segments rather than individual time steps. LLMDiff, a diffusion model using frozen LLM (large language model) Transformers, was built on Earthformer but extended it into a diffusion-based probabilistic setting\cite{She2024}. Bi et al. proposed a novel deep generative framework, combining a Vector Quantized Generative Adversarial Network (VQGAN) and an autoregressive Transformer to produce radar-based precipitation forecasts.\cite{Bi2023}. The model introduces an Extreme Value Loss (EVL) function derived from Extreme Value Theory (EVT) to model rare, high-intensity rainfall events. Liu et al. applied a model named Multi-Level Transformer Fusion (MLTF) to quantitative precipitation estimation in the source region of the Yellow River basin (SRYRB), which is a sparsely-gauged area\cite{LIU2024}. To provide a clearer problem-oriented analysis, Table~\ref{Transformer_comparison} summarizes the contributions of representative Transformer-based models in addressing specific challenges in precipitation prediction.

\begin{table*}[t]
\caption{Brief comparison of Transformer-based models in precipitation prediction}
\label{Transformer_comparison}
\centering
\small
\renewcommand{\arraystretch}{1.25}
\setlength\tabcolsep{0.9pt} 
\begin{tabular}{%
L{0.14\textwidth}
L{0.14\textwidth}
L{0.1\textwidth}
L{0.2\textwidth}
L{0.2\textwidth}
L{0.195\textwidth}}
\hline
\textbf{Nerual Network} &\textbf{Prediction Target} & \textbf{Data} & \textbf{Benchmarks} &\textbf{Best Score} & \textbf{Contribution} \\
\hline
spatiotemporal transformer-based conditional diffusion model\cite{Zhao2024} & Next 1 h & SEVIR & ConvLSTM, MotionRNN, SimVP, SVG, conditional GAN & Best in HSS (0.2142) and CSI (0.1293) under strong rainfall conditions (>32.2 kg/$m^2$) & Integrated a spatiotemporal Transformer within a diffusion framework, enabling global modeling of precipitation evolution\\
\hline
Earthformer \cite{Gao2022} & Next 1 h & SEVIR & Persistence , UNet, ConvLSTM, PredRNN, PhyDNet, E3D-LSTM, Rainformer & 3.6957 (MSE) and 0.4419 (CSI-M) & Provided a state-of-the-art Transformer model using cuboid attention\\
\hline
LLMDiff (frozen LLM Transformer blocks) \cite{She2024} & Next 1 h &  SEVIR & UNet, ConvLSTM, PredRNN, PhyDNet, E3D-LSTM, Rainformer, Earthformer & 3.5581 (MSE) and 0.4508 (CSI-M) & Introduced a transformer block from LLMs to collect long-term temporal context information\\
\hline
LPT-QPN \cite{Li2023} & Next 100 minutes & SEVIR & UNet, DGMR, ConvLSTM, TrajGRU, PredRNNv2, AFNOnet & 0.5363 in CSI-74 and 0.3131 in CSI-133 (heavy rain) & Provided a lightweight physics-informed Transformer enabling longer lead-time and high-intensity areas prediction\\
\hline
MDTNet \cite{Zhaozewei2024} & Short-term (next few hours) & SEVIR / SRAD2018 & ConvLSTM, PFST-LSTM, Rainformer, MIMO-VP, rainymotion, DGMR, STRPM & 0.4009 (HSS), 0.2667 (CSI), 0.3869 (POD) ($\ge12.14$, heavy rain) & Introduced multiscale deformable Transformer, Fourier space regularization loss and adversarial loss\\
\hline
Motion-guided global–local aggregation Transformer \cite{Dong2022} & Next 1 h & SEVIR / SRAD2018 & ConvLSTM, PredRNN, Conv-TT-LSTM, IDA-LSTM, U-Net, rainymotion & 0.3602 (CSI), 0.5009 (HSS) ($\ge3.5kg/m^2$) & Combined short-term motion and long-term context using motion-guided attention \\
\hline
SwinNowcast \cite{LiZhuang2025} & Next 30 minutes (6 frames) & KNMI & UNet, ConvLSTM, PhyDNet, SmaAt-UNet, PredRNN, Rainformer &  0.7494 (CSI), 0.3868 (HSS), 0.1574 (FAR) (lower thresholds, 0.5mm/h) & Introduced 3 modules (M-FBM, MSCBAM and GAFFU) to integrate multi-scale local and global features\\
\hline
HTLA \cite{LiWenhui2024} & Next 45 minutes (9 frames) & KNMI & ConvLSTM, PredRNN++, SmaAt-UNet, AA-TransUNet, Rainformer & $0.6737 \pm 0.0027$ (HSS) and $0.3435 \pm 0.0009$ (CSI) when $r>0.5$ & Reduced Transformer computational burden while improving radar nowcasting\\
\hline
Preformer \cite{Jin2024} & Next 6 hours & ERA5 & ConvLSTM, ConvGRU, TrajGRU, PredRNN, PFST, SCCN, SimVP, SimVP+gSTA, TAU &  88.57 (RL level), 52.54 (LR level) in IoU & Developed a encoder–translator–decoder Transformer, and achieved SOTA performance with fewer parameters\\
\hline
\end{tabular}
\end{table*}

\subsection{Hybrid Neural Networks}
Although single neural network models have significantly improved the precipitation prediction capabilities at multiple scales, future development requires the integration of method innovation with practical applications. The current trend is that hybrid architectures are widely used in precipitation prediction tasks, for they can enhance models' abilities in extracting spatiotemporal information from different resources, and combining physical process knowledge with neural network flexibility can improve both skill and interpretability. 

\begin{figure}
\centering
\includegraphics[width=1\linewidth]{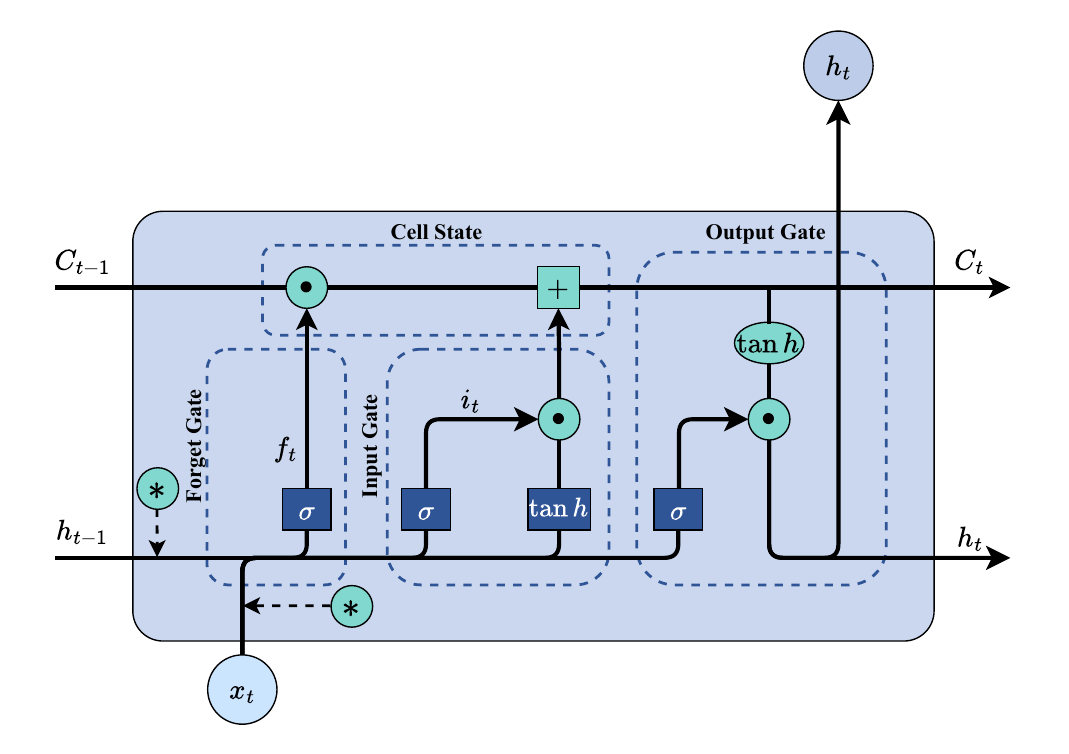}
\caption{Structure of a ConvLSTM cell unit. Unlike standard LSTM, matrix multiplications are replaced by convolutional operations in ConvLSTM.}
\label{ConvLSTM Model}
\end{figure}

An architecture widely used in this academic field is convolutional LSTM (ConvLSTM). In 2015, Shi et al. proposed this pioneering model, specifically designed for spatiotemporal sequence forecasting tasks like precipitation nowcasting\cite{SHI2015}. This model extends traditional fully connected LSTM (FC-LSTM) by integrating convolutional operations into both the input-to-state and state-to-state transitions (shown in Figure~\ref{ConvLSTM Model}), so the transition equations change to:
\begin{align*}
i_t &= \sigma\left( W_{xi} * X_t \;+\; W_{hi} * H_{t-1} \;+\; W_{ci} \circ C_{t-1} \;+\; b_i \right), \\
f_t &= \sigma\left( W_{xf} * X_t \;+\; W_{hf} * H_{t-1} \;+\; W_{cf} \circ C_{t-1} \;+\; b_f \right), \\
\tilde{C}_t &= \tanh\left( W_{xc} * X_t \;+\; W_{hc} * H_{t-1} \;+\; b_c \right), \\
C_t &= f_t \circ C_{t-1} \;+\; i_t \circ \tilde{C}_t, \\
o_t &= \sigma\left( W_{xo} * X_t \;+\; W_{ho} * H_{t-1} \;+\; W_{co} \circ C_t \;+\; b_o \right), \\
H_t &= o_t \circ \tanh(C_t),
\end{align*}

where $*$ and $\circ$ denote the convolution operator and the Hadamard product, respectively. The $i$, $f$ and $o$ represent the input, forget and output gates of the corresponding timestamps. $H$, $C$, and $X$ denote the hidden state, the cell outputs, and the inputs, respectively.
The results of their research show that it outperforms the state-of-the-art ROVER algorithm and FC-LSTM models in several key evaluation metrics (e.g., MSE, CSI, FAR, POD, Corr.). 
Years later, baseline models, Trajectory Gated Recurrent Unit (TrajGRU)\cite{shi2017deep} and PredRNN\cite{WangNIPS2017} were developed as an extension of ConvLSTM to address its limitations in modeling non-stationary and location-variant spatiotemporal dynamics. Whereas ConvLSTM applies fixed convolutional kernels for state transitions across spatial grids, TrajGRU learns location-variant recurrent connections through dynamically generated flow fields. Ignited by these groundbreaking studies, PredRNN++ \cite{wang2018predrnn++} was soon proposed to refine the architecture with Causal LSTMs and a more effective spatiotemporal memory trajectory design. Unlike ConvLSTM’s fixed convolutional recurrence and TrajGRU’s learned but localized trajectories, PredRNN++ emphasized long-term dependency modeling through a flexible memory passing scheme across layers and time steps. To demonstrate the strength of PredRNN++, Bonnet et al. applied this model to nowcast the precipitation in São Paulo with weather radar images\cite{Bonnet2020}. The model shows significant improvement above the conventional precipitation nowcasting model across multiple lead times. Luo et al. also proposed PFST-LSTM, which extended ConvLSTM with a pseudoflow prediction mechanism to explicitly account for the motion of precipitation systems\cite{Luo2021}. Compared with the original ConvLSTM, PFST-LSTM showed enhanced stability and accuracy, particularly in forecasting complex storm dynamics and longer lead times. Gamboa-Villafruela et al. proposed the CNN-LSTM for short-term prediction using the IMERG dataset\cite{GamboaVillafruela2021}. To further exhibit the potential of ConvLSTM architecture in precipitation prediction, Xiong et al. proposed the Contextual Sa-Attention ConvLSTM (CSAConvLSTM) framework\cite{Xiong2021}. This framework does not employ a single mechanism but combines two fundamentally different enhancement methods. The first is the contextual convolution operation between the input data and the network outputs to learn interactive representations. The second is the self-attention operation applied directly to the hidden states to capture local and global features. This specialized design allows the model to better preserve fine spatial appearances and structural details during the rapid evolution of radar echoes. Experimental comparisons demonstrated that CSAConvLSTM outperformed established baseline models in the HKO-7 dataset. Ultimately, it achieved higher accuracy in short-term rainfall forecasts, exhibiting sharper structural preservation of strong radar echo intensities and reduced error accumulation across longer lead times. The Multiscale LSTM model with Attention Mechanism (MLSTM-AM)\cite{TAO2021}, which combines ATWT, LSTM, and attention mechanism to optimize the whole prediction structure, showing more skillful than any single component in the framework. Similarly, Niu et al. improved short-term forecasting by integrating radar echoes, temperature fields, and precipitation data into a hybrid multi-channel ConvLSTM-3D CNN framework, utilizing a balanced loss function to address the uneven distribution of precipitation rates.\cite{Niu2020}. Ebtehaj and Bonakdari \cite{Ebtehaj2024} demonstrated that CNNs offer efficiency and precision in short-term heavy rainfall forecasting, whereas LSTMs achieve superior accuracy for longer lead times and moderate precipitation events. These findings indicate that the hybrid framework, by leveraging CNN for rapid feature extraction and LSTM for sequence-dependent modeling, can provide more robust and practically valuable solutions for real-time flood forecasting and short-term precipitation forecasting. Moreover, Shen et al. combined the fully adaptive noise ensemble empirical modal decomposition (CEEMDAN) algorithm, SVM, and an LSTM to construct a hybrid CEEMDAN-SVM-LSTM model for forecasting monthly precipitation in Lanzhou City\cite{Shen2023}. In their baseline comparisons, they found that while the LSTM outperformed statistical methods like autoregressive integrated moving average (ARIMA), it was actually inferior to the SVM in error metrics such as RMSE, MSE, and MAE. To maximize the strengths of both algorithms, they introduced the CEEMDAN module. By comparing the CEEMDAN-LSTM and CEEMDAN-SVM models with their single-architecture counterparts, they demonstrated that the decomposition module behaved exactly as expected. Finally, the combined CEEMDAN-SVM-LSTM model effectively leveraged the specific advantages of both architectures, yielding a 57.5$\%$ improvement in RMSE, a 55.1$\%$ improvement in MAE, and 33.8$\%$ progress in $R^2$ index compared to the CEEMDAN-LSTM model alone.

Extending this idea, more hybrid models have been developed to enhance LSTM’s predictive capability under non-stationary and highly variable rainfall conditions. By combining wavelet decomposition for denoising, ARIMA for linear dynamics, and LSTM for nonlinear learning, the Wavelet-ARIMA-LSTM model improved robustness across multiple meteorological stations\cite{Wu2021}. Similarly, by utilizing complementary ensemble empirical modal decomposition (CEEMD) to break down rainfall signals into intrinsic components before applying an LSTM, the CEEMD-LSTM model improved accuracy and stability in capturing not only short-term fluctuations but also long-term patterns\cite{Zhang2021}. 

Recently, hybrid models have integrated diverse paradigms to address different forecasting scales. The TelNet model \cite{Pinheiro2025} exemplifies this with a variable selection network to rank feature importance for seasonal precipitation forecasting. Its design demonstrates how interpretable modules can be embedded into modern architectures to enhance trustworthiness while maintaining high predictive performance. Furthermore, the Rainformer model \cite{Bai2022} provides a balanced fusion of local and global features based on two practical components, namely the global features extraction unit and the gate fusion unit. This modular design significantly improve accuracy in high-resolution nowcasting. NowcastNet\cite{Zhang2023NowcastNet}, designed for extreme precipitation prediction, integrates a physics-guided evolution network that explicitly models precipitation advection and intensity evolution alongside a generative refinement network. The evolution component produces motion and growth patterns that correspond to physically meaningful precipitation dynamics, offering a degree of interpretability absent in purely data-driven architectures. 

In addition to task-specific hybrid precipitation models, recent large-scale systems have adopted hybrid architectural designs that combine multiple neural paradigms and operational forecasting constraints. For example, the ECMWF's Artificial Intelligence Forecasting System (AIFS) integrates a GNN encoder–decoder with a Transformer-based processor core for global atmospheric forecasting\cite{lang2024aifs}. The model was trained to produce 6-hour forecasts of multi-level and surface variables, including total and convective precipitation. Precipitation contributes directly to the multi-variable training loss, and forecast skill is evaluated using standard precipitation verification metrics such as stable equitable error in probability space (SEEPS). Other large-scale systems predict precipitation through different structural designs. FourCastNet explicitly forecasts total precipitation by utilizing a secondary diagnostic neural network that processes the predictive outputs of its main atmospheric backbone\cite{pathak2022fourcastnet}. Conversely, while the foundational Pangu-Weather model did not include precipitation among its investigated variables, its three-dimensional Earth-specific Transformer (3DEST) architecture demonstrates great potential and could be applied to complex rainfall forecasting applications in the future.\cite{Bi2023Pangu}.Such systems illustrate how modern hybrid neural networks are evolving toward unified atmospheric prediction frameworks in which precipitation forecasting is embedded within joint dynamical state prediction.

Hybrid models are primarily developed to overcome the structural limitations of single-architecture networks in practical applications by integrating complementary modeling strengths. To summarize, we list representative hybrid models with the specific challenges they address in Table~\ref{Hybrids_comparison}. Due to the powerful capabilities of hybrid models, hybrid neural networks have been increasingly applied to precipitation prediction in recent years. To avoid redundancy, we will not elaborate further here. The relevant articles published recently (2022-2024) are supplemented in Appendix 2.

\begin{table*}[t]
\caption{Brief comparison of Hybrid models in precipitation prediction}
\label{Hybrids_comparison}
\centering
\small
\renewcommand{\arraystretch}{1.25}
\setlength\tabcolsep{0.9pt} 
\begin{tabular}{%
L{0.14\textwidth}
L{0.14\textwidth}
L{0.15\textwidth}
L{0.15\textwidth}
L{0.2\textwidth}
L{0.195\textwidth}}
\hline
\textbf{Nerual Network} &\textbf{Prediction Target} & \textbf{Data} & \textbf{Benchmarks} & \textbf{Best Score} & \textbf{Contribution} \\
\hline
ConvLSTM \cite{SHI2015} & Next 90 minutes (15 frames) & Weather radar intensities in Hong Kong &  ROVER algorithm, FC-LSTM & 1.420 (MSE), 0.577 (CSI), 0.195 (FAR), 0.660 (POD), 0.908 (Correlation) & First convolutional recurrent model \\
\hline
PredRNN \cite{WangNIPS2017}& Next 60 minutes (10 frames) &Radar echo dataset & ConvLSTM, PixelCNN-based VPN & 44.2 (MSE/frame) & Unified spatiotemporal memory cell (zigzag memory flow) \\
\hline
PredRNN++ \cite{Bonnet2020} & 15-150 minutes (10 steps) & Local weather radar data & ENCAST & Statistical comparison in selected cases & Validate PredRNN++ (Causal LSTM + Gradient Highway) in regional radar data \\
\hline
CEEMDAN-SVM-LSTM \cite{Shen2023} & Monthly & Local station's rainfall data & LSTM, ARIMA, BP, SVM, SGBOOST & 7.68 (RMSE), 58.98 (MSE), 5.58 (MAE), 0.95 ($R^2$) & Combine signal decomposition with machine learning method\\
\hline
\end{tabular}
\end{table*}
\section{Discussion}
\subsection{Review of Neural Networks}
\begin{figure}
    \centering
    \includegraphics[width=1\linewidth]{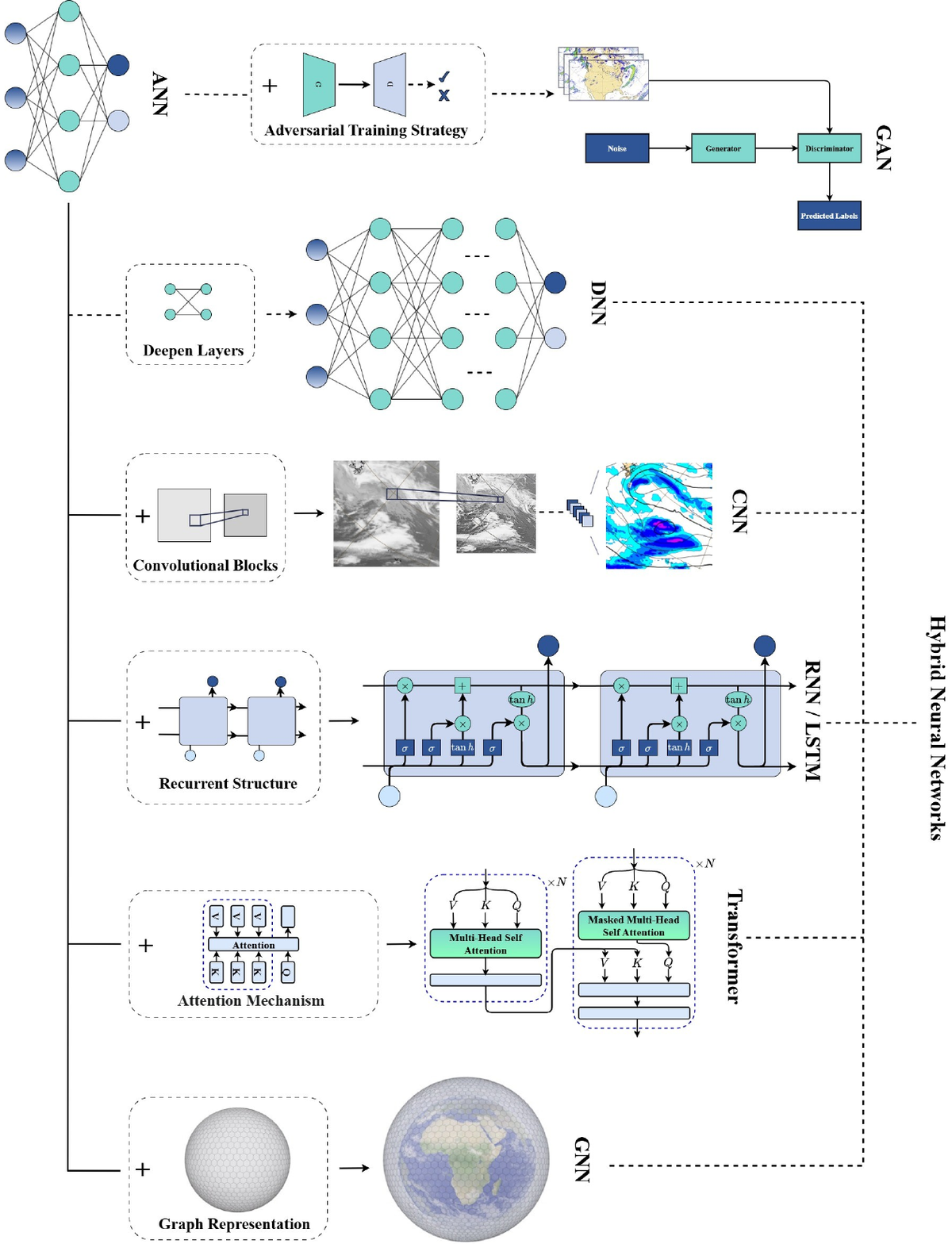}
    \caption{Connections and differences between neural networks. All neural networks evolved from ANN and are distinguished from each other through specific architectures. For instance, the recurrent structure of RNN/LSTM. It is worth noting that GAN does not introduce any specific architecture but incorporates the adversarial training method.}
    \label{Connections}
\end{figure}

This article gives a comprehensive review of the evolution of neural network architectures in precipitation prediction, ranging from early ANN-based networks to contemporary hybrid models. Table~\ref{ANNDNN_comparison} lists some representative ANN/DNN-based models and highlights their methodological contributions in precipitation prediction. It can be summarized that the structure of ANNs is simple (such as 3 or 4 layers), and the compared baselines are mostly regression methods and statistical methods. However, these simple neural networks have demonstrated potential in the field of precipitation prediction, paving the way for the current applications of complex neural networks. DNN is an advanced version of ANN, by increasing the depth of the hidden layers, to enhance the ability to represent high-dimensional data. The main advantage of the ANN/DNN model lies in its simple concept and relatively lower computational requirements compared to modern deep architectures. The disadvantages are also obvious. They lack explicit processing capabilities for spatial or temporal information. Table~\ref{CNN_comparison} compares demonstrative CNN-based precipitation prediction models. By replacing dense connections with convolutional kernels, CNN models explicitly encode spatial features, enabling effective extraction of organized rainfall structures. Compared with ANN-based approaches, CNN models demonstrate clear improvements in high-resolution radar-based nowcasting and satellite-driven rainfall forecasting. Moreover, CNN-based prediction models are not limited to short-term predictions. They also show potential in monthly precipitation prediction by learning spatial patterns related to circulation. Despite these advantages, CNN models mainly focus on local spatial dependence and can not model temporal evolution. Table~\ref{RNNLSTM_comparison} presents representative RNN/LSTM-based models, which extend neural network-based forecasting into the temporal dimension. Unlike CNNs that primarily capture spatial features, recurrent architectures incorporate the memory mechanism and can simulate the sequential evolution of rainfall. This ability to obtain temporal information is particularly important for short-term continuous precipitation prediction, where precipitation at the following time step strongly depends on prior atmospheric states. The gating mechanism in LSTMs alleviates the problem of gradient vanishing and enables the network to retain relevant historical information over a longer time range. Additionally, the multimodal RNN framework further enhances prediction performance by integrating radar, atmospheric variables, and satellite-derived features. However, RNN and LSTM fundamentally rely on sequential processing, which limits parallelization and increases computational latency for long sequences. Table~\ref{GNN_comparison} lists 2 representative GNN models. GNN models operate on graph representations, enabling flexible message passing across irregular spatial nodes and multi-scale connections. By integrating physical drivers, physics-informed GNN demonstrates superior performance in maintaining regional precipitation variability compared to traditional NWP and convolutional baselines. Although GNN-based systems have many advantages compared to CNNs, the construction of graphs, the design of adjacency relationships, and scalability remain significant challenges. Table~\ref{GAN_comparison} summarizes GAN-based precipitation prediction models. Unlike conventional approaches that optimize pixel-wise reconstruction losses, GANs introduce adversarial learning to encourage sharper, and more realistic precipitation structures. By learning the underlying data distribution rather than minimizing only pixel-wise errors, GAN-based models better preserve high-intensity rainfall cores and fine-scale convective structures. This property is practical for extreme precipitation forecasting, where regression-based models tend to underestimate peak intensities due to averaging effects. Table~\ref{Transformer_comparison} summarizes Transformer-based precipitation prediction models. Transformer achieves flexible long-range spatiotemporal interaction across grid cells or image blocks by replacing fixed convolution operations and recurrent memory modules with self-attention operations. This property is beneficial under strong convective conditions or large-domain forecasting scenarios, where precipitation evolution depends on nonlocal atmospheric interactions. The multiple studies listed in the table show that in terms of the CSI, HSS and RMSE indicators, the Transformer-based models always outperform ConvLSTM, the CNN-based encoder-decoder framework and traditional statistical methods. Although these Transformer models have strong predictive performance and scalability, they usually require a large amount of training data and computing resources. Unless an optimized attention mechanism is adopted, their quadratic attention complexity may be unmanageable for high-resolution radar sequences. Moreover, the interpretability of attention weights does not necessarily translate into physical interpretability, which remains a challenging research problem that has not yet been solved. Table~\ref{Hybrids_comparison} presents representative hybrid neural network architectures that integrate multiple modeling approaches to better capture the multi-scale and nonlinear characteristics of the precipitation process. Hybrid models typically combine convolutional layers for spatial feature extraction, recurrent or attention modules for temporal evolution, and in some cases, also incorporate statistical decomposition or physical constraints to enhance physical consistency. Compared with a single architecture model, the hybrid framework demonstrates superior  prediction capabilities, especially in the scenario of radar-based short-term forecasting. Architectures such as ConvLSTM and PredRNN integrate spatial and temporal memory mechanisms, enabling precipitation evolution to be modeled within a single recurrent structure. However, the hybrid model still faces the risks of high computational cost and overfitting, especially when there are limited samples of extreme events. As the number of structural components increases, the interpretability of the model may also decrease. Nevertheless, hybrid neural networks are mainstream in short-term and medium-term precipitation forecasting research, which reflects the trend of multi-scale and multimodal modeling strategies. Finally, to provide an intuitive and consolidated comparison of major neural network architectures applied to precipitation prediction, we show their main difference in Figure~\ref{Connections} and list their advantages and disadvantages in Table~\ref{NNs_comparison}.

\begin{table*}[t]
\caption{Comparison of neural networks in precipitation prediction studies}
\label{NNs_comparison}
\centering
\small
\renewcommand{\arraystretch}{1.25}
\setlength\tabcolsep{1pt} 
\begin{tabular}{%
L{0.15\textwidth}
L{0.28\textwidth}
L{0.28\textwidth}
L{0.27\textwidth}}
\hline
\textbf{Nerual Network} & \textbf{Strengths} & \textbf{Limitations} & \textbf{Typical Tasks} \\
\hline
ANN / DNN & Simple regression with classification & No spatial modeling & Station rainfall prediction \\
\hline
CNN & Captures spatial rainfall structures & Limited long-range temporal modeling & Radar-based forecasting; Satellite precipitation mapping \\
\hline
RNN / LSTM & Temporal sequence modeling & Training instability; Slow sequential computation & Time-series rainfall prediction \\
\hline
GAN & Generates realistic rainfall fields & Hard to evaluate physically & High-resolution rainfall generation \\
\hline
Transformer & Long-range spatiotemporal dependency modeling & High computational cost; Large training data required & Global forecasting \\
\hline
GNN & Natural modeling on spherical grids & Complex implementation & Global NWP emulation \\
\hline
Hybrid NN & Better capture multi-scale spatiotemporal rainfall evolution & High architectural complexity; Risk of overfitting with limited extreme-event data & Multimodal precipitation fusion \\
\hline
\end{tabular}
\end{table*}

Apart from large models mentioned in the last section, which thoroughly adopt neural networks as core architectures, some neural network–based infrastructures established by meteorological institutions also indicate a gradual transition of weather prediction from research-oriented studies toward operational evaluation and potential forecasting applications. For example, community initiatives supported by NOAA and academic partners now provide large reforecast archives and visualization platforms enabling comparisons between AI-based forecasting systems such as FourCastNet, Pangu-Weather and GraphCast with operational forecasting models\cite{Jacob2025}. Complementary community efforts of NOAA further emphasize the development of research-to-operations pathways, including evaluation infrastructure and operational readiness planning for AI-based forecasting systems\cite{Frolov2025}. These developments demonstrate that neural networks are increasingly being assessed within operational forecasting workflows, where improvements in precipitation prediction represent a key application target.
It is worth noting that many recent deep learning models predict future radar echo fields rather than rainfall directly. Precipitation forecasts are subsequently obtained through radar–rainfall conversion or QPE modules. Models such as the Recurrent Evolution Memory-Aware Network (REMNet)\cite{Jing2022}, the Memory in Memory (MIM)\cite{MIM2019}, the Multisource Data Model (MSDM)\cite{MSDM2021}, the Temporal and Spatial GAN (TSGAN)\cite{ChenTSGAN2022}, the two-stage UA-GAN\cite{Xu2022}, and the Nonlocal Echo Dynamics (NLED) network\cite{NLED2022} follow this paradigm and demonstrate strong performance in short-term nowcasting. While models like REMNet, MIM, and NLED are technically echo extrapolation frameworks that implicitly provide precipitation forecasts via standard external conversions, systems like MSDM actively embed machine learning QPE modules (such as RF) directly into their architecture to explicitly estimate rainfall rates. Together, these methods represent an important component of modern precipitation prediction systems.

\subsection{Existing Problems and Future Directions}
During the past decade, neural networks have made significant progress in precipitation prediction. However, there are still some problems. Based on a comprehensive review of existing studies, we outline the most pressing challenges currently facing the field and identify the corresponding key directions for future research required to achieve reliable and operational predictions. 

\subsubsection{Generalization and Region-Agnostic Modeling}
Many existing models are often trained and evaluated within certain geographical areas, resulting in poor performance when applied to regions with different precipitation patterns. To address this issue, future research should focus on developing region-independent frameworks, such as through transfer learning, domain adaptation, or pre-training on global climate datasets, followed by targeted fine-tuning. This will enable models trained in local areas to effectively transfer to larger regions, thereby enhancing their global applicability.

\subsubsection{Representation of Extreme Events and Optimization of Loss Function}
The precipitation dataset suffers from a severe imbalance in extreme samples, as heavy rainfall events occur infrequently but are the main source of hydrological risks. As discussed in the previous sections of this review, since standard loss functions underestimate these high-impact rainfall amounts, researchers have developed various loss functions. Therefore, future neural network designs must be tailored to specific tasks and precipitation distribution to design loss functions, thereby balancing overall predictive ability with better detection and intensity estimation of rare events. 

\subsubsection{Mitigation of Data Scarcity}
Unlike short-term weather datasets, long-term precipitation records have a small amount of data and are of poor quality. As mentioned earlier, even a widely used large dataset like ERA5 has sparse observation data in the early period of the Southern Hemisphere. Moreover, since datasets are calibrated typically based on frequent and moderate weather conditions, the accuracy of extreme precipitation in the records is relatively low. Additionally, due to factors such as sparse rain gauge networks, radar beam blocking effects, and difficulties in satellite inversion, precipitation observations in complex terrains still have significant uncertainties. When using highly parameterized and data-demanding models to train based on these limited datasets, the models are prone to overfitting. To address the issue of specific data scarcity, future research needs to preprocess such data before neural network training, and expand the training dataset reasonably according to the characteristics of precipitation distribution.

\subsubsection{Enhancement of Explainability and Physics-Informed Integration} 
Purely data-driven neural networks usually have poor interpretability. Although some methods can improve the interpretability of the model, they usually describe statistical correlations rather than true causal mechanisms. This lack of physical basis often raises concerns about the reliability of predictions in complex or unprecedented weather conditions. To bridge this gap, future research should increasingly integrate heterogeneous data sources such as radar, satellite images, reanalysis data, and climate indices into physics-informed neural networks or hybrid statistical frameworks. By explicitly constraining the neural network architecture with physical laws of atmospheric processes (such as Navier-Stokes Equations and Conservation of Mass), the model can utilize complementary information from multiple observation platforms while enhancing physical consistency. Ultimately, combining advanced neural networks with physical modeling components will enable predictions to capture actual precipitation dynamics, thereby significantly improving the interpretability and reliability of the system.

\subsubsection{Improvement of Practicality and Standardization of Benchmarks} 
Some advanced and complex models, such as the Transformers and GNNs, are capable of capturing long-range spatiotemporal dependencies. However, they usually require a significant amount of computational resources for training and inference. Due to these memory and computational limitations, existing models typically struggle to effectively perform global high-resolution precipitation predictions, thereby limiting their practicality. The development of scalable architectures that can dynamically adjust spatial resolution or computational focus is crucial for the future neural networks.

Finally, through the comparison in this article, it is found that even within the same type of neural networks, due to the inconsistency of indicators and datasets, it is difficult to conduct a complete evaluation for each neural network. Establishing standardized benchmarks and open evaluation datasets will help in conducting highly transparent comparisons between models, improving the reproducibility of research results, and conducting strict evaluations of the true generalization performance of the models.

{\appendix
\section*{Appendix 1: List of Abbreviations}
CBAM: convolutional block attention module 

DWD: German Weather Service

GPM: Global Precipitation Measurement

NCEI: National Centers for Environmental Information

AWS: automatic weather station

WGAN: Wasserstein generative adversarial network

STPF: short-term precipitation forecast

MMF: multimodal fusion

KMA: Korea Meteorological Administration

ALPF: advanced lightweight precipitation forecasting

ATWT: à trous (French term meaning “with holes”) wavelet transform

CMA: China Meteorological Administration

GLDAS: Global Land Data Assimilation System

GEFS: Global Ensemble Forecast System

FMI: Finnish Meteorological Institute 

FMCGEP: fuzzy control-based multicellular gene expression programming

RRDBNet: Residual-in-Residual Dense Block based Network

SCSSA: Sparrow optimization algorithm incorporating positive cosine and Cauchy variants

BILSTM: bi-directional LSTM

WSR-88D: Weather Surveillance Radar-1988 Doppler

NASA: National Aeronautics and Space Administration

JAXA: Japan Aerospace Exploration Agency

pySTEP: An open-source Python library for probabilistic precipitation nowcasting\cite{Pulkkinen2019}

TIGGE: THORPEX Interactive Grand Global Ensemble

HREF: High-Resolution Ensemble Forecast

MLR: Multiple Linear Regression

BP: Back Propagation

SVM: Support Vector Machine

RF: Random Forest

M-FBM: multi-scale feature balancing module

MSCBAM: multi-scale convolutional block attention module

GAFFU: gated attention feature fusion unit

ROVER: Real-time Optical flow by Variational methods for Echoes of Radar

PFST-LSTM: pseudo flow spatiotemporal LSTM

CRPS: continuous ranked probability score
\section*{Appendix 2: List of Supplement References}
Supplement references are listed in Table X.

\section*{Appendix 3: List of Evaluation Metrics}
Evaluation metrics are listed in Table XI.
}

\clearpage  

\begingroup
\renewcommand{\arraystretch}{1.5}
\setlength\tabcolsep{1.2pt} 
\onecolumn
\begin{longtable}{L{0.33\textwidth}L{0.1\textwidth}L{0.15\textwidth}L{0.14\textwidth}L{0.15\textwidth}L{0.1\textwidth}}

\hline
\toprule
\textbf{Dataset} &\textbf{Temporal Range} &\textbf{Spatial Area} &\textbf{Model Type} &\textbf{Evaluation Metrics} & \textbf{ref}\\
\midrule
\endfirsthead

\multicolumn{6}{l}%
{\tablename\ \thetable\ -- \textit{Continued from previous page}} \\
\hline
\toprule
\textbf{Dataset} & \textbf{Temporal Range} &\textbf{Spatial Area} & \textbf{Model Type}  & \textbf{Evaluation Metrics} & \textbf{ref}\\

\endhead

\hline \multicolumn{6}{l}{\textit{Continued on next page}} \\ \hline
\endfoot

\bottomrule
\endlastfoot


PRISM/West-WRF dataset & 1985--2019  & Western United States & U-Net CNN & RMSE, MAE, BIAS, CSI, PC & Badrinath et al.\cite{Badrinath2023} \\
\hline
Atmospheric variables from ERA5/Rain gauge data from CMA & 1960--2015 & Central-Eastern China & MLP-CNN & Accuracy, Recall, AUC, $S$ & Jiang et al.\cite{JIANG2024}\\
\hline
Grid data from ECMWF/Meteorological data from local stations & 2013--2019 & Longzhong Loess Plateau region, China & CNN-LSTM & Accuracy, Precision, Recall, $F_1$, CSI  & Li et al.\cite{LiWeide2022}\\
\hline
Meteorological observation data from China Ground Cumulative Daily Value Dataset (V3.0) & 1953--2019 & Kunming, China & PCA-CNN-BiLSTM-Attention & NSE, MAE, RMSE, CC, $\alpha$, $\beta$ & Guo et al. \cite{Guo2025}\\
\hline
Precipitation data from ERA5 & 1979--2019 & China & CBAM-CNN & PCC, RMSE & Jin et al.\cite{Jin2022}\\
\hline
Fengyun 4A Satellite Data/Precipitation data from IMERG & 2018--2021& Southeast coast of China & Attention-Unet & POD, CSI, FAR, RMSE, CC & Gao et al.\cite{GaoYanbo2022}\\
\hline
Radar data from FMI & 2019--2021 & The Baltics and Finland & Lagrangian CNN (L-CNN) & POD, FAR, ETS, FSS, MAE, ME & Ritvanen et al.\cite{Ritvanen2023}\\
\hline
GEFS v2/ERA5& 1985--2019 & Sacramento River Basin, United States & CNN combined with MLP & F1 Score, AUC & Zhang et al.\cite{Zhangchen2022}\\
\hline

Data from ERA5 and IMERG & 2001--2020 & The middle reaches of the Yellow River, China & RRDBNet (CNN based) & RMSE, CC, percentage bias & Fu et al.\cite{Fu2024}\\
\hline
HKO-7/DWD-12/MeteoNet/Brasil & 2009--2015/2006--2017/2016--2018/2015--2019 & HongKong, China/Germany/ France/S\~{a}o Paulo, Brazil & PrecipLSTM & HSS, CSI & Ma et al.\cite{Ma2022}\\
\hline
ERA5, IMERG, and GLDAS & 2013--2017 & Qinghai Province, China & 3D CNN and bidirectional ConvLSTM & MAE, RMSE, CC & You et al.\cite{You2025}\\
\hline
Data from local meteorological stations & 1961--2020 & Changde, China  & CEEMD-PSO-LSTM & RMSE, MAE, MAPE & X. Jiang\cite{Jiang2023}\\
\hline
Precipitation data provided by the authors& 1951--2021 & Ankang City, and Yongding District, China & VMD-MSMA-LSTM-ARIMA & RMSE, MAPE, MAE, $R^2$, NSE, PICP, MPI & Cui et al.\cite{Cui2023}\\
\hline
Data of various local rainfall stations & 2000--2019 & Luoyang City, China & EMD-VMD-LSTM & RMSE, $\delta$ & Guo et al.\cite{Guo2023}\\
\hline
Data collected by ground weather stations& 2018--2019 & North China & 3D-SA-LSTM & CSI, HSS, FAR, POD & Chen et al.\cite{Chen2022}\\
\hline
Tropical Cyclone Precipitation Dataset (TCPD) & 2017--2019 & Northwest Pacific & spatiotemporal Graph-guided Convolutional LSTM & MSE, MAE & Yang et al.\cite{YANG2022}\\
\hline
Monthly precipitation data from NCEI & 1973--2021 & Luoyang City, China & EEMD-LSTM-ARIMA & RMSE, MAE, MSE, $R^2$ & Zhao et al.\cite{Zhao2022}\\
\hline
Real-world radar echo map dataset provided by the authors& 2014--2018 & Guangdong province, China & Multiscale WGAN & HSS, CSI, POD, FAR & Luo et al.\cite{Luo2022}\\
\hline
Radar reflectivity data/Precipitation data from regional AWSs provided by the authors& May to September in 2017 and 2018 & Southeastern region of China & STPF-Net & CSI, POD, FAR, BIAS, HSS, BMSE, BMAE & Wang et al.\cite{Wang2024}\\
\hline
Radar echo data from the CIKM AnalytiCup 2017& A total time span of 3 years & Multiple sites  & PredRANN & MSE, B-MSE, SSIM, HSS, CSI & Luo et al.\cite{PredRANNLUO2022}\\
\hline
RAIN-F dataset from KMA & 2017--2019 & Korean Peninsula & MMF-RNN & CSI, HSS, PSNR, B-MAE, B-MSE & Liu et al. \cite{Liu2025}\\
\hline
SEVIR dataset & 2017--2019 & Contiguous US & ALPF (CNN based) & RMSE, SSIM, PSNR, LPIPS & Yang et al.\cite{Yang2024}\\
\hline
WSR-88D radar data/local rain gauge data & 2016--2019 & Florida Peninsula, United States & RQPENet & RMSE, MAE, CC, NSE, BIAS, POD, FAR, CSI, HSS & Li et al.\cite{LiWenyuan2023}\\
\hline
Measured monthly rainfall data from the authors & 1996--2020 & Xi'an City, China & SCSSA-CNN-BILSTM & $R^2$, RMSE, MAE & Zhang et al.\cite{ZhangXianqi2024}

\label{Supplement papers}
\end{longtable}

\endgroup
  {{\begin{center}
      \text{Table X:} \text{Supplement references for this review.}
  \end{center}}}
\clearpage  

\clearpage  

\begingroup
\renewcommand{\arraystretch}{1.5}
\setlength\tabcolsep{1.2pt} 
\onecolumn
\begin{longtable}
{|L{0.29\textwidth}|C{0.4\textwidth}|L{0.295\textwidth}|}

\hline
\textbf{Metric} & \textbf{Formula} & \textbf{Description} \\ \hline
\endfirsthead

\multicolumn{3}{l}%
{\tablename\ \thetable\ -- \textit{Continued from previous page}} \\
\hline
\textbf{Metric} & \textbf{Formula} & \textbf{Description}  \\ \hline
\endhead

\hline \multicolumn{3}{l}{\textit{Continued on next page}} \\ \hline
\endfoot

\hline
\endlastfoot

RMSE (Root Mean Square Error) & 
$\displaystyle \sqrt{\frac{1}{N}\sum_{i=1}^{N}(P_i - O_i)^2}$ & 
Overall magnitude of prediction errors. Here, $P_i$ = predicted value, $O_i$ = observed value, $N$ = total samples.\\ 
\hline
MSE (Mean Square Error) & 
$\displaystyle \frac{1}{N}\sum_{i=1}^{N}(P_i - O_i)^2$ & 
Average squared error. Here, $P_i$ = predicted value, $O_i$ = observed value, $N$ = total samples.\\ \hline

MAE (Mean Absolute Error) & 
$\displaystyle \frac{1}{N}\sum_{i=1}^{N}\left|P_i - O_i\right|$ & 
Average absolute error magnitude. Here, $P_i$ = predicted value, $O_i$ = observed value, $N$ = total samples.\\ \hline

PRE & 
$\displaystyle\frac{\mathrm{MAE}(p_i,\hat{p}_i)}{\bar{p}_i} $ & 
The proportional regional error for region \(i\). Here, $p_i$ is the benchmark rainfall series, $\hat{p}_i$ is the predicted rainfall series, and $\bar{p}_i$ is the mean benchmark rainfall for region \(i\), respectively.\\ \hline

PME &
$\displaystyle\frac{\mathrm{MAE}(p_m,\hat{p}_m)}{\bar{p}_m} $ & 
The proportional monthly error for month \(m\). Here, $p_m$ is the benchmark rainfall series, $\hat{p}_m$ is the predicted rainfall series, and $\bar{p}_m$ is the mean benchmark rainfall for month \(m\), respectively.\\ \hline

ME (Mean Error) & 
$\displaystyle \frac{1}{N}\sum_{i=1}^{N}(P_i - O_i)$ & 
Average bias of predictions. Here, $P_i$ = predicted value, $O_i$ = observed value, $N$ = total samples.\\ \hline

B-MAE &
$\displaystyle \frac{1}{C} \sum_{c=1}^{C} \frac{1}{n_c} \sum_{i=1}^{n_c} \left| P_i^{(c)} - O_i^{(c)} \right|$ &
Balanced Mean Absolute Error, here \(C\) is the number of classes; \(n_c\) is the number of samples in class \(c\); \(P_i^{(c)}\) and \(O_i^{(c)}\) denote the predicted and observed values, respectively, for the \(i\)-th sample in class\(c\).
\\ \hline

B-MSE &
$\displaystyle \frac{1}{C} \sum_{c=1}^{C} \frac{1}{n_c} \sum_{i=1}^{n_c} \left( P_i^{(c)} - O_i^{(c)} \right)^2$ &
Balanced Mean Squared Error, here \(C\) is the number of classes; \(n_c\) is the number of samples in class \(c\); \(P_i^{(c)}\) and \(O_i^{(c)}\) denote the predicted and observed values, respectively, for the \(i\)-th sample in class\(c\).\\
\hline

FSS (Fraction Skill Score)&
$\displaystyle 1- \frac{\frac{1}{N}\sum_{i=1}^{N}(P_i - O_i)^2}{\frac{1}{N}\sum_{i=1}^{N}({P_i}^2 + {O_i}^2)}$ &
The Fraction Skill Score evaluates the spatial agreement between predicted 
and observed precipitation fields after neighborhood smoothing. Here, $P_i$ = predicted value, $O_i$ = observed value, $N$ = total samples.\\
\hline

MAPE (Mean Absolute Percentage Error) & 
$\displaystyle \frac{100\%}{N} \sum_{i=1}^{N} \left| \frac{P_i - O_i}{O_i} \right|$ & 
Average absolute percentage difference between predicted and actual values. Here, $P_i$ = predicted value, $O_i$ = observed value, $N$ = total samples.\\ \hline

$R^2$ (Coefficient of Determination) & 
$\displaystyle 1 - \frac{\sum_{i=1}^{N} (P_i - O_i)^2}{\sum_{i=1}^{N} (O_i - \bar{O})^2}$ & 
Proportion of variance in observed data explained by the predictions.Here, $P_i$ = predicted value, $O_i$ = observed value, $N$ = total samples.\\ \hline

BIAS & 
$\displaystyle \frac{\sum_{i=1}^{N} P_i}{\sum_{i=1}^{N} O_i}$ & 
Overestimation/underestimation. Here, $P_i$ = predicted value, $O_i$ = observed value, $N$ = total samples.\\ \hline

Relative Bias & 
$\displaystyle \frac{\sum_{i=1}^{N} P_i}{\sum_{i=1}^{N} O_i} - 1$ & 
Relative overestimation/underestimation. Here, $P_i$ = predicted value, $O_i$ = observed value, $N$ = total samples.\\ \hline

CSI (Critical Success Index) & 
$\displaystyle \frac{H}{H + M + F}$ & 
Fraction of observed events correctly predicted. Here, $H$ = correctly predicted (hits), $M$ = misses, $F$ = false alarms.\\ \hline

PC (Probability of Correct) & 
$\displaystyle \frac{H + C}{N}$ & 
Overall accuracy (hits + correct negatives). Here, $H$ = correctly predicted (hits), $C$ = correct negatives, $N$ = total samples.\\ \hline

CC (Correlation Coefficient) & 
$\displaystyle \frac{\sum_{i=1}^{N} (P_i-\bar{P})(O_i-\bar{O})}{\sqrt{\sum_{i=1}^{N} (P_i-\bar{P})^2 \sum_{i=1}^{N} (O_i-\bar{O})^2}}$ & 
Linear association between prediction and observation. Here, $P_i$ = predicted value, $O_i$ = observed value, $\bar{P}$ = mean of all predicted values, $\bar{O}$ = mean of all observed values, $N$ = total samples.\\ \hline

POD (Probability of Detection) & 
$\displaystyle \frac{H}{H + M}$ & 
Fraction of observed events that were predicted. Here, $H$ = correctly predicted (hits), $M$ = misses.\\ \hline

POFD (Probability of False Detection) & 
$\displaystyle \frac{F}{F + C}$ & 
Fraction of non-events incorrectly predicted as events. Here, $F$ = false alarms, $C$ = correct negatives.\\ \hline

FAR (False Alarm Ratio) & 
$\displaystyle \frac{F}{H + F}$ & 
Fraction of predicted events that did not occur. Here, $F$ = false alarms, $H$ = correctly predicted (hits).\\ \hline

SR (SuccessRatio) &
$\displaystyle \frac{H}{H + F}$ & 
Fraction of the forecasted precipitation events that were correctly observed. Here, $F$ = false alarms, $H$ = correctly predicted (hits).\\ \hline

HSS (Heidke Skill Score) & 
$\displaystyle \frac{2(HC - MF)}{(H+M)(M+C) + (H+F)(F+C)}$ & 
Skill score adjusted for random chance. Here, $H$ = hits, $M$ = misses, $F$ = false alarms, $C$ = correct negatives.\\ \hline

NSE (Nash-Sutcliffe Efficiency) & 
$\displaystyle 1 - \frac{\sum_{i=1}^{N} (O_i - P_i)^2}{\sum_{i=1}^{N} (O_i - \bar{O})^2}$ & 
Predictive skill vs. mean observation. Here, $P_i$ = predicted value, $O_i$ = observed value, $N$ = number of samples.\\ \hline

Accuracy &
$\displaystyle \frac{TN + TP}{n} \times 100\%$ &
Percentage of all samples correctly classified. Here, $TN$ = true negatives, $TP$ = true positives, $n$ = total number of predictions.\\ \hline

$S$ (Positive Rank Sum) &
$\displaystyle  \sum_{i \in \text{positives}} \mathrm{rank}_i$ &
Sum of the ranks assigned to positive instances when all samples are ranked by score. \\ \hline

AUC (Area Under ROC Curve) &
$\displaystyle \frac{S - \frac{P(P+1)}{2}}{N \times P}$ &
Probability that a randomly chosen positive ranks higher than a randomly chosen negative. Here, $S$ is Positive Rank Sum, and P(N) represent the number of positive (negative) events.\\ \hline

Recall (Sensitivity) &
$\displaystyle \frac{TP}{TP + FN} \times 100\%$ &
Fraction of actual positives that are correctly identified. Here, $TP$ = true positives, $FN$ = false negatives.\\ \hline

Precision &
$\displaystyle \frac{TP}{TP + FP} \times 100\%$ &
Fraction of predicted positives that are actual positives. Here, $TP$ = true positives, $FP$ = false positives.\\ \hline

$F_1$ &
$\displaystyle 2 \times \frac{\text{Precision} \times \text{Recall}}{\text{Precision} + \text{Recall}}$ &
Harmonic mean of Precision and Recall (denoted above), balancing both metrics. \\ \hline

PBIAS (Percent Bias/Water Volume Error) &
$\displaystyle  100\times\frac{\sum_{i=1}^N (P_i - O_i)}{\sum_{i=1}^N O_i}$ &
Percentage bias of predicted volume relative to the observed; positive values indicate overestimation, while negative values indicate underestimation. Here, $P_i$ = predicted value, $O_i$ = observed value, $N$ = number of samples.\\ \hline

$\alpha$ & 
$\displaystyle \frac{\sigma_p}{\sigma_o}$ &  
$\sigma_p$ and $\sigma_o$ are the standard values of predicted and simulated precipitation, respectively\\ \hline

$\beta$ & 
$\displaystyle \frac{{y}_{p,i} - {y}_{o,i}}{{y}_{o,i}}$ &  
${y}_{p,i}$ and ${y}_{o,i}$ are the simulated value of precipitation and observed value of precipitation at time $i$, respectively\\ \hline

KGE (Kling-Gupta Efficiency) &
$\displaystyle 1 - \sqrt{(CC - 1)^2 + \left(\frac{\sigma_P }{\sigma_O } - 1\right)^2 + \left(\frac{\mu_P}{\mu_O} - 1\right)^2}$ &
Composite metric combining correlation $CC$, variability ratio, and bias ratio to assess agreement between predictions and observations. Here, $\sigma_P$ and $\sigma_O$ are standard deviations for predictions and observations, respectively. $\mu_P$ and $\mu_O$ are means of predictions and observations, respectively. $CC$ is the Correlation Coefficient denoted above.\\ \hline

mKGE (modified Kling-Gupta Efficiency) &
$\displaystyle 1 - \sqrt{(CC - 1)^2 + \left(\frac{\sigma_P / \mu_P}{\sigma_O / \mu_O} - 1\right)^2 + \left(\frac{\mu_P}{\mu_O} - 1\right)^2}$ &
A modified version of $KGE$, which ensures that the bias and variability ratios are not cross-correlated. Here, $\sigma_P$ and $\sigma_O$ are standard deviations for predictions and observations, respectively. $\mu_P$ and $\mu_O$ are means of predictions and observations, respectively. $CC$ is the Correlation Coefficient denoted above.\\ \hline

OE (Overestimation) &
$\displaystyle HPB+FB$ &
Total positive bias from HPB and FB. HPB and FB are components from 4CED\cite{Zhang2021GL094092}\\ \hline

UE (Underestimation) &
$\displaystyle HNB + MB$ &
Total negative bias from HNB and MB. HNB and MB are components from 4CED\cite{Zhang2021GL094092}\\ \hline

ETS/GSS &
$\displaystyle \frac{TP - R}{TP + FP + FN - R}$ &
The Equitable Threat Score (ETS)/Gilbert skill score (GSS) measures forecast skill relative to random chance. Here, $TP$ = true positives, $FP$ = false positives, $FN$ = false negatives, $R= \frac{(TP + FP)(TP + FN)}{\text{Total}}$\\ \hline

PICP (Prediction Interval Coverage Probability)& 
$\displaystyle \displaystyle  \frac{1}{n} \sum_{i=1}^{n} c_i$ & 
Measures the proportion of ground truth values that fall within their corresponding prediction intervals. Here, $c_i=1$ is the judgement coefficient if $P_i$ lies in the prediction interval, otherwise $c_i=0$.\\ \hline

MPI (Mean Prediction Interval) &
$\displaystyle 2 \cdot t_{n-p}^{\alpha/2} \cdot \left( \frac{1}{n} \sum_{i=1}^{n} s_i \right)$ &
Measures the average width of the prediction intervals across all samples. The coefficient $t_{n-p}^{\alpha/2}$ is the $t$-distribution with $n-p$ degrees of freedom for $(1-\alpha)\%$.\\ \hline

$\delta$ (Actual Relative Error) &
$\displaystyle \frac{\Delta}{L}\times100\%$ &
Refers to the value obtained by multiplying the absolute error $\Delta$ caused by the measurement by the ratio of the measured (conventional) true value$L$, and then multiplying by $100\%$, expressed in percentage.\\ \hline

PSNR (Peak Signal-to-Noise Ratio) &
$\displaystyle 10 \cdot \log_{10} \left( \frac{MAX_I^2}{\text{MSE}} \right)$ &
Where $MAX_I$ is the maximum possible pixel value of the image (for example, $MAX_I = 255$ for 8-bit grayscale images). Higher PSNR values indicate greater similarity (less distortion).
\\ \hline

SSIM (Structural Similarity Index) &
$\displaystyle \frac{ (2 \mu_x \mu_y + C_1)(2 \sigma_{xy} + C_2) }
     { (\mu_x^2 + \mu_y^2 + C_1)(\sigma_x^2 + \sigma_y^2 + C_2) }$ &
Where $\mu_x, \mu_y$ are the mean intensities of $x$ and $y$, $\sigma_x^2, \sigma_y^2$ are their variances, $\sigma_{xy}$ is the covariance between $x$ and $y$, $C_1, C_2$ are small constants to stabilize the division. SSIM values range from $-1$ to $1$. A value of $1$ indicates perfect similarity, while values close to $0$ or negative indicate poor structural similarity.\\ \hline

LPIPS (Structural Similarity Index) &
$\displaystyle \sum_{l} \frac{1}{H_l W_l} 
\sum_{h,w}
\left\|(\hat{y}_l(x)_{hw} - \hat{y}_l(y)_{hw}) \right\|_2^2 $ &
Where $\hat{y}_l(x)$ and $\hat{y}_l(y)$ are unit-normalized feature maps from layer $l$ of the network, $H_l, W_l$ are spatial dimensions of the feature map at layer $l$, $w_l$ are learned weights that calibrate the contribution of each channel. LPIPS has been shown to correlate strongly with human perceptual studies. Lower LPIPS score indicates higher perceptual similarity.
\\ \hline

Correlation &
$\displaystyle \frac{\sum_{i,j} P_{i,j} \, T_{i,j}}
{\sqrt{\left(\sum_{i,j} P_{i,j}^{2}\right)
\left(\sum_{i,j} T_{i,j}^{2}\right)} + \varepsilon} $ &
Where $\varepsilon$ is a small constant added for numerical stability. $i$, $j$ is the row/column index of the image. $P$ is the predicted frame, and $T$ is a ground-truth frame. \\ \hline

Brier Score (BS) &
$\displaystyle \frac{1}{N} \sum_{i=1}^{N} \left( p_i - o_i \right)^2$&
Where $p_i$ is the forecast probability of the event, $o_i$ is the corresponding observation. $N$ is the total number of forecast cases. \\ \hline

CRPS & 
$\displaystyle \int_{-\infty}^{\infty}
\left[ F(y) - F_o(y) \right]^2 \, dy $ &
Where $F_o(y) =
\begin{cases}
0, & y < \text{observed value}, \\
1, & y \geq \text{observed value}.\end{cases}$, $F(y)$ is approximated using the ensemble of forecasts 
generated by a given model.

\label{tab:evaluation_metrics}
\end{longtable}
\endgroup

  {{\begin{center}
      \text{Table XI:} \text{Evaluation metrics for neural network precipitation models.}
  \end{center}}
}

\clearpage  

\bibliographystyle{unsrtnat}
\bibliography{references}

\end{document}